\pdfoutput=1

\documentclass{article}
\usepackage{colm2024_conference}

\usepackage{url}
\usepackage{booktabs}
\usepackage{graphicx}
\usepackage{microtype}
\usepackage{multicol}
\usepackage{xcolor}
\usepackage{multirow}
\usepackage{lipsum}
\usepackage{enumitem}
\usepackage{pifont}
\usepackage{colortbl}
\usepackage{tabularx}
\usepackage{caption} 
\usepackage{float}
\usepackage{amsmath,amsfonts,amssymb,bbm}
\usepackage{adjustbox}
\usepackage{makecell}
\usepackage{arydshln}
\usepackage{fvextra}
\usepackage[breakable]{tcolorbox}
\usepackage{subcaption}
\usepackage{hyperref}

\newcommand*\samethanks[1][\value{footnote}]{\footnotemark[#1]}

\title{The Lessons of Developing Process Reward Models\\in Mathematical Reasoning}

\author{
\textbf{Zhenru Zhang \quad Chujie Zheng \quad Yangzhen Wu \quad Beichen Zhang \quad Runji Lin} \\
\textbf{Bowen Yu\thanks{Corresponding authors.} \quad Dayiheng Liu\samethanks{} \quad Jingren Zhou \quad Junyang Lin\samethanks{} } \\
\vspace{0.15cm}
Qwen Team, Alibaba Group
}

\begin{document}
\maketitle

\begin{abstract}
Process Reward Models (PRMs) emerge as a promising approach for process supervision in mathematical reasoning of Large Language Models (LLMs), which aim to identify and mitigate intermediate errors in the reasoning processes.
However, the development of effective PRMs faces significant challenges, particularly in data annotation and evaluation methodologies.
In this paper, through extensive experiments, we demonstrate that commonly used Monte Carlo (MC) estimation-based data synthesis for PRMs typically yields inferior performance and generalization compared to LLM-as-a-judge and human annotation methods.
MC estimation relies on completion models to evaluate current-step correctness, which can generate correct answers from incorrect steps or incorrect answers from correct steps, leading to inaccurate step verification.
Furthermore, we identify potential biases in conventional Best-of-N (BoN) evaluation strategies for PRMs:
(1) The unreliable policy models generate responses with correct answers but flawed processes, leading to a misalignment between the evaluation criteria of BoN and the PRM objectives of process verification. 
(2) The tolerance of PRMs of such responses leads to inflated BoN scores.
(3) Existing PRMs have a significant proportion of minimum scores concentrated on the final answer steps, revealing the shift from process to outcome-based assessment in BoN Optimized PRMs.
To address these challenges, we develop a consensus filtering mechanism that effectively integrates MC estimation with LLM-as-a-judge and advocates a more comprehensive evaluation framework that combines response-level and step-level metrics. Based on the mechanisms, we significantly improve both model performance and data efficiency in the BoN evaluation and the step-wise error identification task.
Finally, we release a new state-of-the-art PRM that outperforms existing open-source alternatives and provides practical guidelines for future research in building process supervision models.


\begin{figure}[hb]
  \centering
  \includegraphics[width=0.99\linewidth]{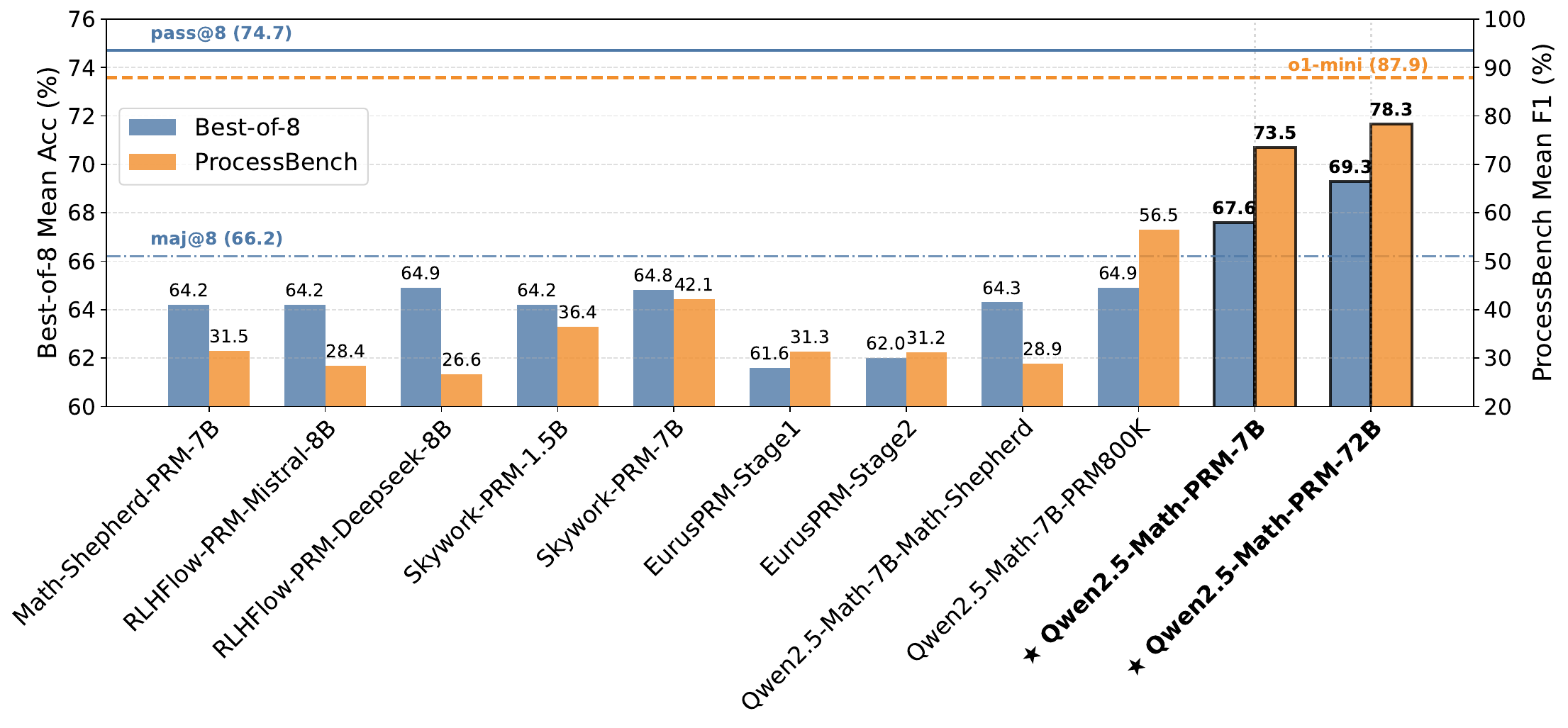}
  \caption{Overview of evaluation results on the Best-of-8 strategy of the policy model Qwen2.5-Math-7B-Instruct and the benchmark \textsc{ProcessBench} \citep{processbench} across multiple PRMs (see Table \ref{tab:best-of-n on Qwen2.5-Math-7b-Instruct} and Table \ref{tab:processbench} for details).}
  \label{fig:overall_res}
\end{figure}

\end{abstract}

\section{Introduction}

In recent years, Large Language Models (LLMs) have made remarkable advances in mathematical reasoning \citep{gpt4,llama3,deepseek-math,deepseek-coder-v2,qwen2,qwen2.5-math,qwen2.5}, yet they can make mistakes, such as miscalculations or logical errors, leading to wrong conclusions.
Moreover, even when achieving correct final answers, these powerful models can still regularly make up plausible reasoning steps, where the final answers build upon flawed calculations or derivations, which undermine the reliability and trustworthiness of LLMs' reasoning processes. 
To address these challenges, Process Reward Models (PRMs; \citealt{prm, math-shepherd}), as a representative and recently focal approach, are proposed to identify and mitigate process errors, thereby enabling finer-grained supervision on the reasoning process.

One critical challenge of developing PRMs lies in the data annotation for the correctness of reasoning processes, which is typically expensive and time-consuming. 
While \citet{prm} recruited human annotators with detailed instructions and elaborate procedures to achieve satisfactory annotation quality, the prohibitive cost pushes researchers to explore automated annotation methods. 
Among them, one commonly used approach is to assess process correctness by estimating the empirical probability of leading to the correct final answers through Monte Carlo (MC) methods, which has attracted great research interests and has also been commonly employed in practice \citep{rlhflow-prm,math-shepherd, luo2024improvemathematicalreasoninglanguage}. Another challenge lies in evaluating PRM performance, as previous studies \citep{prm, math-shepherd,luo2024improvemathematicalreasoninglanguage} have predominantly relied on the Best-of-N (BoN) evaluation, which selects the highest-scored response from $N$ candidates according to a PRM. Recently, \textsc{ProcessBench} \citep{processbench} have emerged to evaluate the capability of PRMs in identifying step-wise correctness.

Nevertheless, during the training of our own PRM following conventional principles to construct data using MC estimation and evaluate on BoN, we gain several crucial lessons. \textbf{In terms of MC estimation}, (1) we observe that the PRM trained via MC estimation demonstrated significantly inferior performance and generalization capabilities compared to LLM-as-a-judge \citep{llm-as-a-judge} and human annotation. (2) We attribute the suboptimal performance of MC estimation to its fundamental limitation, which attempts to evaluate deterministic current-step correctness based on potential future outcomes. It significantly relies on the performance of the completion model, which may generate correct answers based on incorrect steps, or incorrect answers based on correct steps, introducing substantial noise and inaccuracy verification into step-wise correctness estimation. \textbf{Regarding the BoN evaluation}, (1) the unreliable policy models generate responses with correct answers but flawed processes, leading to a misalignment between the evaluation criteria of BoN and the PRM objectives of process verification. (2) the limited process verification capability makes PRMs demonstrate tolerance for these cases, resulting in inflated BoN performance. (3) We find that in the step scores distribution of existing PRMs, a significant proportion of minimum scores are concentrated on the final answer steps, indicating PRMs have shifted from process to outcome-based assessment in BoN.

To address these challenges, we develop a consensus filtering mechanism that combines MC estimation with LLM-as-a-judge. The instances are only retained when both LLM-as-a-judge and MC estimation show consensus on the error reasoning step locations in the solution. Our approach demonstrates more efficient data utilization and surpass existing open-source PRMs in the conventional BoN evaluation. Furthermore, we advocate for complementing response-level BoN with step-wise evaluation methods. We employ the step-wise benchmark \textsc{ProcessBench} \citep{processbench} to measure the ability to identify process errors in mathematical reasoning. Our trained PRMs exhibit impressively stronger error identification performance than other open-source models, from PRMs to general language models, confirming that our training approach genuinely teaches PRMs to assess the correctness of intermediate reasoning steps.

Our key contributions can be summarized as follows:
\begin{itemize}
\item We identify critical limitations in current data construction approaches for PRMs, demonstrating that MC estimation-based data construction yields inferior performance compared to LLM-as-a-judge and human annotation.
\item We reveal the potential bias in using response-level BoN evaluation alone for PRMs and advocate for comprehensive evaluation strategies combining both response-level and step-level metrics.
\item We propose a simple yet efficient consensus filtering mechanism that integrates MC estimation with LLM-as-a-judge, significantly improving both model performance and data efficiency in PRM training.
\item We substantiate our findings through extensive empirical studies and also open source our trained PRMs, which can establish practical guidelines and best practices for future research and development for reasoning process supervision.
\end{itemize}

\section{Preliminary Trials}\label{Preliminary Trials}
In this section, we describe our preliminary attempts to train PRMs via MC estimation-based reasoning step annotation.
Despite our efforts in scaling up training data and careful tuning of training objectives, we found that the MC estimation-based PRMs do not possess noticeable advantages over the one trained on human-annotated data \citep{prm}, and even lag significantly behind the latter in identifying specific erroneous reasoning steps.

\subsection{Training Setup}\label{sec:Training Setup}

\paragraph{Training Data Synthesis}
We followed the commonly used MC estimation approach, Math-Shepherd \citep{math-shepherd}, to construct the PRM training data.
Specifically, we collected a large-scale dataset of approximately 500,000 queries with golden answers. 
For each query, we generate 6-8 diverse responses by mixing outputs from the Qwen2-Math-Instruct and Qwen2.5-Math-Instruct series models \citep{qwen2.5-math}, spanning the model sizes of 7B and 72B parameters. 
These responses are systematically split into individual steps using the delimiter ``\textbackslash n\textbackslash n''. 
To assess the correctness of each step, we conduct 8 independent completions starting from this step using Qwen2.5-Math-Instruct series with the corresponding model size, estimating the step labels based on the empirical probabilities of each step yielding the correct final answer.
We trained PRMs with either hard labels or soft labels. For \textit{hard} labels, we treat a step as correct if any one of the 8 completions yields the correct final answer, and negative otherwise. 
For \textit{soft} labels, we determined the value (between 0 and 1) as the proportion of completions leading to the correct final answers.
Note that we eliminated all steps subsequent to those labeled as incorrect (label 0), as their validity becomes irrelevant after an error occurs. This removal was implemented to prevent potential model confusion during training.

\paragraph{Training Details}
Our trained PRMs were initialized from the supervised fine-tuned Qwen2.5-Math-7B/72B-Instruct models \citep{qwen2.5-math}, where we replace the original language modeling head (used for next token prediction) with a scalar-value head, consisting of two linear layers.
We calculated the cross-entropy (CE) loss and mean squared error (MSE) loss on the last tokens of each step for the binary classification task using hard labels and for the regression task using soft labels, respectively.

\subsection{Evaluation Setup}\label{sec:Evaluation Setup}

We evaluate our trained PRMs from two aspects: their utilities in straightforwardly improving downstream task performance and their abilities to identify specific erroneous steps in reasoning processes.

\paragraph{Best-of-N}
Consistent with previous work \citep{prm,math-shepherd,luo2024improvemathematicalreasoninglanguage,gsm8k,qwen2.5-math}, we employed the Best-of-N (BoN) sampling strategy for evaluation, which selects the highest-scored response from $N$ candidates according to a PRM.
We denote the evaluation metric as ``prm@$N$''.
Following \citet{qwen2.5-math}, we sampled eight responses (i.e., $N=8$) from Qwen2.5-Math-7B-Instruct across multiple mathematical benchmarks, including GSM8K \citep{gsm8k}, MATH \citep{math}, Minerva Math \citep{minerva}, GaoKao 2023 En \citep{mario}, OlympiadBench \citep{olympiadbench}, College Math \citep{mathscale}, and MMLU STEM \citep{mmlu}.
Each candidate response is scored using the product of all the individual scores of each step within the response, as computed in \citet{prm}.
We also report the result of majority voting among eight samplings (maj@8) as the baseline, and pass@8 (i.e., the proportion of test samples where any of the eight samplings lead to the correct final answers) as the upper bound.

\paragraph{\textsc{ProcessBench}} We also evaluated on \textsc{ProcessBench} as a complement. \textsc{ProcessBench} \citep{processbench} measures the capability of models to identify erroneous steps in mathematical reasoning. Models are required to identify the first step that contains an error or conclude that all steps are correct. Following the evaluation methods for PRMs in \textsc{ProcessBench}, we locate the first erroneous step from predict scores yielded by PRMs.

\begin{table}[!t]
\centering
\resizebox{1\textwidth}{!}{
\begin{tabular}{lcccccccc}
\toprule
\textbf{Setting} & \textbf{GSM8K} & \textbf{MATH} & \textbf{\makecell{Minerva\\Math}} & \textbf{\makecell{GaoKao\\2023 En}} & \textbf{\makecell{Olympiad\\Bench}} & \textbf{\makecell{College\\Math}} & \textbf{\makecell{MMLU\\STEM}} & \textbf{Avg.} \\ \midrule
pass@8 (Upper Bound) & 98.1 &	92.0	& 49.3	& 80.5	& 59.6	& 52.6	& 90.5	& 74.7 \\
\midrule
maj@8 & 96.7	& 87.1	& \textbf{41.2}	& \textbf{72.5}	& \textbf{44.4}	& 47.8	& \textbf{73.8}	& \textbf{66.2} \\ 
Qwen2.5-Math-7B-PRM800K & \textbf{96.9} & 86.9	& 37.1	& 71.2	& 44.0	& 47.6	& 70.9	& 64.9 \\
Qwen2.5-Math-7B-PRM-MC-hard  & 96.8	& \textbf{87.3}	& 40.1	& 70.6	& 43.7	& \textbf{48.1} & 71.6	& 65.5 \\
Qwen2.5-Math-7B-PRM-MC-soft  & 96.8	& 86.3	& 37.9	& 70.6	& 41.0	& 47.7	& 70.4	& 64.4 \\
\bottomrule
\end{tabular}
}
\caption{Performance comparison on Best-of-8 using PRMs trained with MC estimated hard labels and soft labels, human-annotated PRM800K, denoted as Qwen2.5-Math-7B-PRM-MC-hard, Qwen2.5-Math-7B-PRM-MC-soft, and Qwen2.5-Math-7B-PRM800K, respectively.}
\label{tab: detail alpha results on Best-of-8}
\end{table}

\begin{table}[!t]
\centering
\resizebox{\textwidth}{!}{
\begin{tabular}{lccccccccccccc}
\toprule
\multirow{2}{*}{\textbf{Model}} & \multicolumn{3}{c}{\textbf{GSM8K}} & \multicolumn{3}{c}{\textbf{MATH}} & \multicolumn{3}{c}{\textbf{OlympiadBench}} & \multicolumn{3}{c}{\textbf{Omni-MATH}} & \multirow{2}{*}{\textbf{Avg. F1}} \\
\cmidrule(lr){2-4} \cmidrule(lr){5-7} \cmidrule(lr){8-10} \cmidrule(lr){11-13}
 & error & correct & \textbf{F1} & error & correct  & \textbf{F1} & error & correct  & \textbf{F1} & error & correct  & \textbf{F1} \\
\midrule
Qwen2.5-Math-7B-PRM800K & 53.1 & 95.3 & 68.2 & 48.0 & 90.1 & \textbf{62.6} & 35.7 & 87.3 & \textbf{50.7} & 29.8 & 86.1 & \textbf{44.3} & \textbf{56.5} \\
Qwen2.5-Math-7B-PRM-MC-hard & 67.1 & 	90.2	& 77.0	& 35.2	& 65.8	& 45.8	& 13.2	& 28.0	& 17.9	& 13.3	& 41.9	& 20.2	& 40.2 \\
Qwen2.5-Math-7B-PRM-MC-soft & 65.7	& 93.3	& \textbf{77.1}	& 35.7	& 64.5	& 46.0	& 13.2	& 29.2	& 18.1	& 12.9	& 40.2	& 19.6	& 40.2 \\
\bottomrule
\end{tabular}
}
\caption{Performance comparison on \textsc{ProcessBench} using PRMs trained with MC estimated hard labels and soft labels, human-annotated PRM800K, denoted as Qwen2.5-Math-7B-PRM-MC-hard, Qwen2.5-Math-7B-PRM-MC-soft, and Qwen2.5-Math-7B-PRM800K, respectively.}
\label{tab: detail alpha results on ProcessBench}
\end{table}

\subsection{Evaluation Results} \label{sec:preliminary evaluation results}
As shown in Table \ref{tab: detail alpha results on Best-of-8} and Table \ref{tab: detail alpha results on ProcessBench}, we denote the models trained on our MC estimated dataset as Qwen2.5-Math-7B-PRM-MC-hard (trained with hard labels) and Qwen2.5-Math-7B-PRM-MC-soft (trained with soft labels), respectively.
To compare them with a baseline model, we trained exclusively on the PRM800K \citep{prm} dataset with its hard labels named Qwen2.5-Math-7B-PRM-PRM800K. 
The experimental results reveal two critical limitations: 
(1) In the Best-of-8 evaluation, none of the PRMs achieved prm@8 scores superior to maj@8. 
(2) When evaluating on the \textsc{ProcessBench} for identifying erroneous reasoning steps, both Qwen2.5-Math-7B-PRM-MC-hard and Qwen2.5-Math-7B-PRM-MC-soft exhibit significantly inferior erroneous step localization capabilities compared to Qwen2.5-Math-7B-PRM-PRM800K, though the former had larger scale of data.

These undesirable evaluation performances push us to reflect on the currently prevalent data synthesis approach and evaluation strategy. Through the subsequent optimization process, we have indeed gained several observations and lessons learned.

\section{The lessons}
In this section, we present the critical lessons gained during the PRM training. Our discussion comprises three main aspects: (1) the limitations of commonly adopted MC estimation approaches in PRMs training, and (2) the bias in using BoN as the sole evaluation metric for optimizing PRMs.

\subsection{Limitations of MC Estimation for PRMs Training}

\subsubsection{Distinguishing PRMs from Value Models}\label{sec:Distinguishing PRMs from Value Models}
Reward models in mathematical reasoning serve as correctness verifiers and PRMs provide fine-grained supervision by evaluating the correctness of intermediate reasoning steps. In contrast, value models estimate the potential of reaching the correct final answer from the current step in the future. The key difference between PRM and value model lies in that PRMs function as deterministic evaluators of current step correctness, while value models operate as predictive estimators of future solution potential.

MC estimation attempts to estimate the potential of reaching the correct final answer in the future from the current step. When we follow this approach to construct data and train the PRMs, the value model principles are incorporated into PRMs training essentially. This methodology potentially introduces performance and generalization limitations which we will discuss in subsequent sections.

\subsubsection{MC Estimation vs. LLM-as-a-judge vs. Human Annotation}\label{sec:MC Estimation vs. LLM-as-a-judge vs. Human Annotation}
We found that MC estimation methods limit PRM's capability to identify erroneous steps as demonstrated in the experiments of Section \ref{sec:preliminary evaluation results}. For further investigation, we compare the performance using 3 distinct data construct approaches: MC estimation, LLM-as-a-judge, and human annotation. 
For the MC estimation approach, we respectively train the PRM on 445k open-source datasets Math-shepherd \citep{math-shepherd} and our 860k similarly constructed dataset. For our constructed dataset, the MC estimation employs responses from Qwen2-Math-Instruct and completes subsequent reasoning processes by Qwen2.5-Math-Instruct. 
For the LLM-as-a-judge approach, we use the same 860k query and response and employ Qwen2.5-72B-Instruct to verify the correctness of each step in the responses with the prompt template shown in Appendix~\ref{sec:llm-as-a-judge prompt}. 
For the human annotation approach, we use the open-source dataset PRM800K \citep{prm} which consists of approximately 265k samples after deduplication against the test set. 

The experimental results of Best-of-8 and \textsc{ProcessBench} are shown in Table \ref{tab: best-of-8 for MC Estimation vs. Human Annotation vs. LLM-as-a-judge} and Table \ref{tab: ProcessBench for MC Estimation vs. Human Annotation vs. LLM-as-a-judge}, respectively. 
For Best-of-8, Table \ref{tab: best-of-8 for MC Estimation vs. Human Annotation vs. LLM-as-a-judge} shows that the PRM trained on our MC estimated data achieves the best average accuracy and human annotation performs worst. 
For \textsc{ProcessBench}, Table \ref{tab: ProcessBench for MC Estimation vs. Human Annotation vs. LLM-as-a-judge} demonstrates that human annotation achieves the best performance with the least amount of data, followed by LLM-as-a-judge, while MC estimation performed the worst despite having the largest dataset overall. 
Specifically, (1) human annotation, despite being only performed on the MATH dataset, exhibited superior generalization capabilities on more complex tasks OlympiadBench and Omni-MATH. 
(2) Given identical data with different annotation approaches, LLM-as-a-judge demonstrates better generalization performance on challenging problems than MC estimation, although the latter showed favorable results on GSM8K. 
(3) For MC estimation, a comparison between our 860k dataset and Math-Shepherd 440k data indicates that performance improvements can still be achieved through data scaling.
The two models trained on MC estimated and human-annotated data exhibit inverse performance relationships in Best-of-8 and \textsc{ProcessBench}, which catches our attention and is thoroughly investigated in Section \ref{sec:Bias in Best-of-N Sampling for PRM Performance Evaluation}.

\begin{table}[!t]
\centering
\resizebox{1\textwidth}{!}{
\begin{tabular}{lccccccccc}
\toprule
\textbf{Setting} & \textbf{\# samples} & \textbf{GSM8K} & \textbf{MATH} & \textbf{\makecell{Minerva\\Math}} & \textbf{\makecell{GaoKao\\2023 En}} & \textbf{\makecell{Olympiad\\Bench}} & \textbf{\makecell{College\\Math}} & \textbf{\makecell{MMLU\\STEM}} & \textbf{Avg.} \\ \midrule
MC Estimation (Math-Shepherd) & 440k & 96.9	& 86.5	& 36.8	& \textbf{71.4} & 	41.6	& 47.7	& 69.3 & 64.3 \\
MC Estimation (our data) & 860k & \textbf{97.0}	& \textbf{87.6}	& \textbf{41.9}	& \textbf{71.4}	& 43.6	& \textbf{48.2}	& \textbf{71.9} & \textbf{65.9} \\
LLM-as-a-judge (our data) & 860k &  96.9 & 	86.8	& 39.0	& 71.2	& 43.7	& 47.7	& \textbf{71.9} & 65.3 \\
Human Annotation (PRM800K) & 264k & 96.9	& 86.9	& 37.1	& 71.2	& \textbf{44.0}	& 47.6	& 70.9 & 64.9 \\
\bottomrule
\end{tabular}
}
\caption{PRMs performance comparison on the Best-of-8 strategy of the policy model Qwen2.5-Math-7B-Instruct. The models are trained on the different data construction methods including MC estimation, LLM-as-a-judge, and human annotation.}
\label{tab: best-of-8 for MC Estimation vs. Human Annotation vs. LLM-as-a-judge}
\end{table}

\begin{table}[!t]
\centering
\resizebox{1\textwidth}{!}{
\begin{tabular}{lcccccccccccccc}
\toprule

\multirow{2}{*}{\textbf{Method}}     & \multirow{2}{*}{\textbf{\# samples}} & \multicolumn{3}{c}{\textbf{GSM8K}} & \multicolumn{3}{c}{\textbf{MATH}} & \multicolumn{3}{c}{\textbf{OlympiadBench}} & \multicolumn{3}{c}{\textbf{Omni-MATH}} & \multirow{2}{*}{\textbf{Avg.F1}} \\

\cmidrule(lr){3-5} \cmidrule(lr){6-8} \cmidrule(lr){9-11} \cmidrule(lr){12-14}
&                            & error  & correct  & \textbf{F1}    & error  & correct  & \textbf{F1}   & error     & correct     & \textbf{F1}       & error    & correct   & \textbf{F1}     &                       \\ \midrule
MC Estimation (Math-Shepherd) & 440k                       & 46.4   & 95.9     & 62.5  & 18.9   & 96.6     & 31.6 & 7.4       & 93.8        & 13.7     & 4.0      & 95.0      & 7.7    & 28.9                  \\
MC Estimation  (our data)     & 860k                       & 62.3   & 91.2     & \textbf{74.0}  & 35.2   & 71.9     & 47.3 & 12.7      & 41.3        & 19.4     & 12.1     & 54.4      & 19.8   & 40.1                  \\
LLM-as-a-judge (our data)   & 860k                       & 44.0   & 99.0     & 60.9  & 33.5   & 94.8     & 49.5 & 24.7      & 97.1        & 39.4     & 22.3     & 95.4      & 36.1   & 46.5                  \\
Human Annotation (PRM800K)  & 264k                       & 53.1   & 95.3     & 68.2  & 48.0   & 90.1     & \textbf{62.6} & 35.7      & 87.3        & \textbf{50.7}     & 29.8     & 86.3      & \textbf{44.3}   & \textbf{56.5}                  \\ \bottomrule
\end{tabular}
}
\caption{PRMs performance comparison on \textsc{ProcessBench}. The models are trained on the different data construction methods including MC estimation, LLM-as-a-judge, and human annotation.}
\label{tab: ProcessBench for MC Estimation vs. Human Annotation vs. LLM-as-a-judge}
\end{table}

\subsubsection{Stringent Data Filtering Mechanisms Required in MC Estimation}\label{sec:Stringent Data Filtering Mechanisms Required in MC Estimation}
We attribute the inferior performance of MC estimation compared to LLM-as-a-judge and human annotation to its high noise in reasoning step correctness estimation and inaccurate error position identification due to its heavy dependence on the policy model. For instance, the policy model may generate correct final answers but incorrect reasoning steps, which will be investigated thoroughly in Section \ref{sec:Unreliable Policy Models Cause BoN-PRMs Misalignment}. 

\begin{figure}[htp]
    \begin{minipage}[t]{0.32\textwidth}
        \centering
        \includegraphics[width=\textwidth]{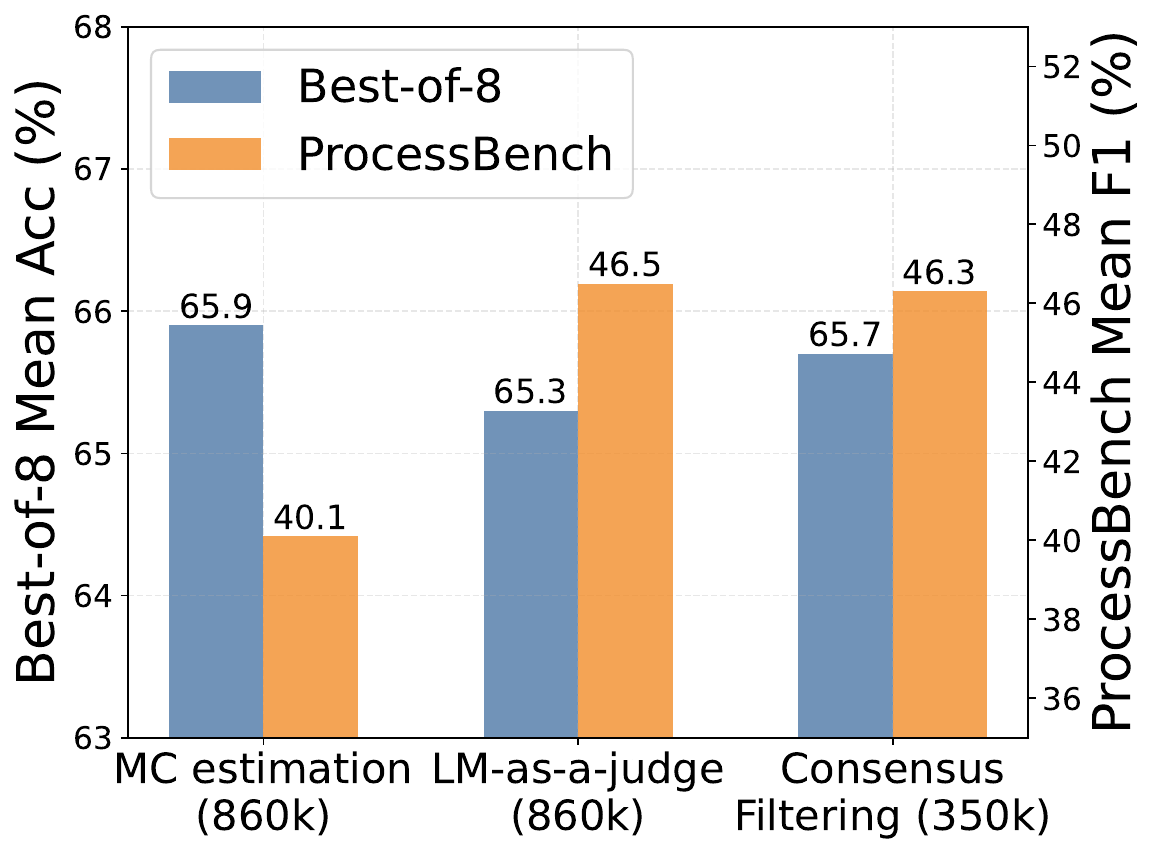}
        \caption{Performance comparison on Best-of-8 and \textsc{ProcessBench} using PRMs trained with different data synthesis methods.}
        \label{fig:consensus_filtering}
    \end{minipage}
    \hfill
    \begin{minipage}[t]{0.32\textwidth}
        \centering
        \includegraphics[width=\textwidth]{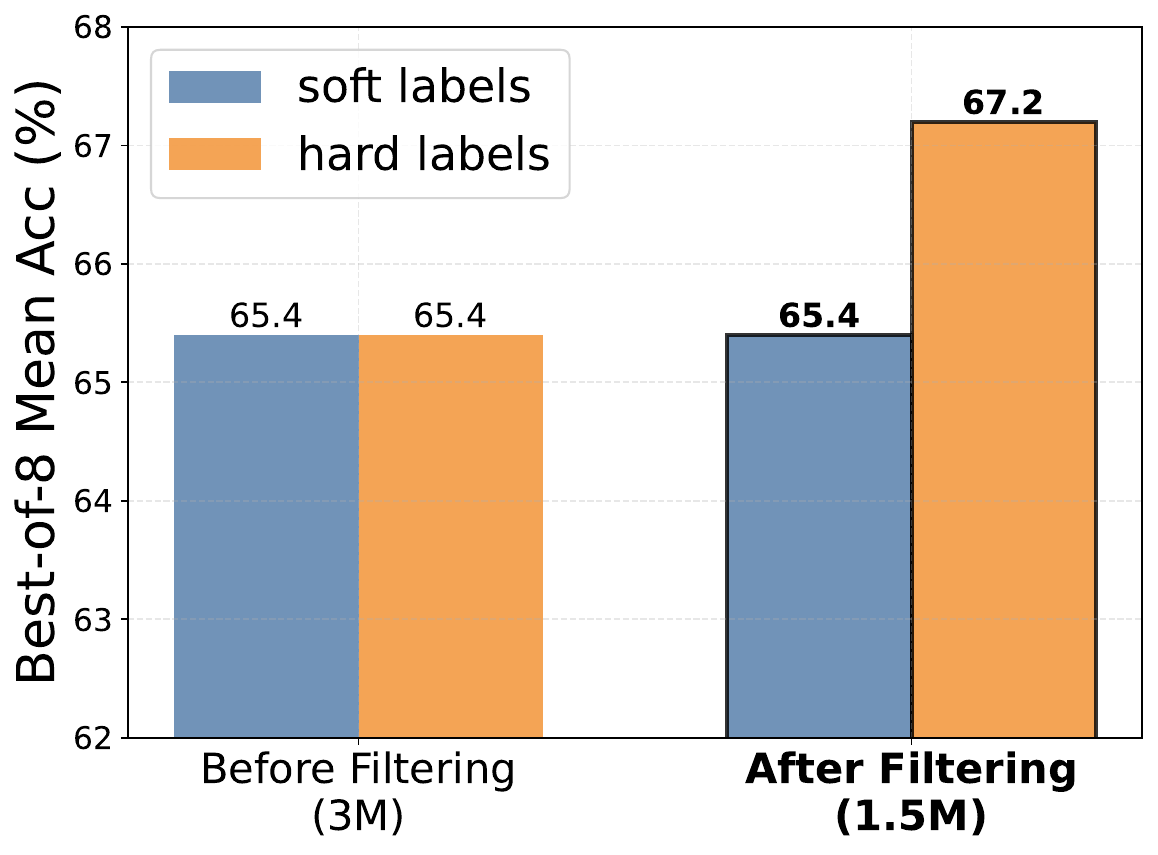}
        \caption{Performance comparison on Best-of-8 for the PRMs trained on soft and hard labels before and after consensus filtering.}
        \label{fig:soft_hard_bon}
    \end{minipage}
    \hfill
    \begin{minipage}[t]{0.32\textwidth}
        \centering
        \includegraphics[width=\textwidth]{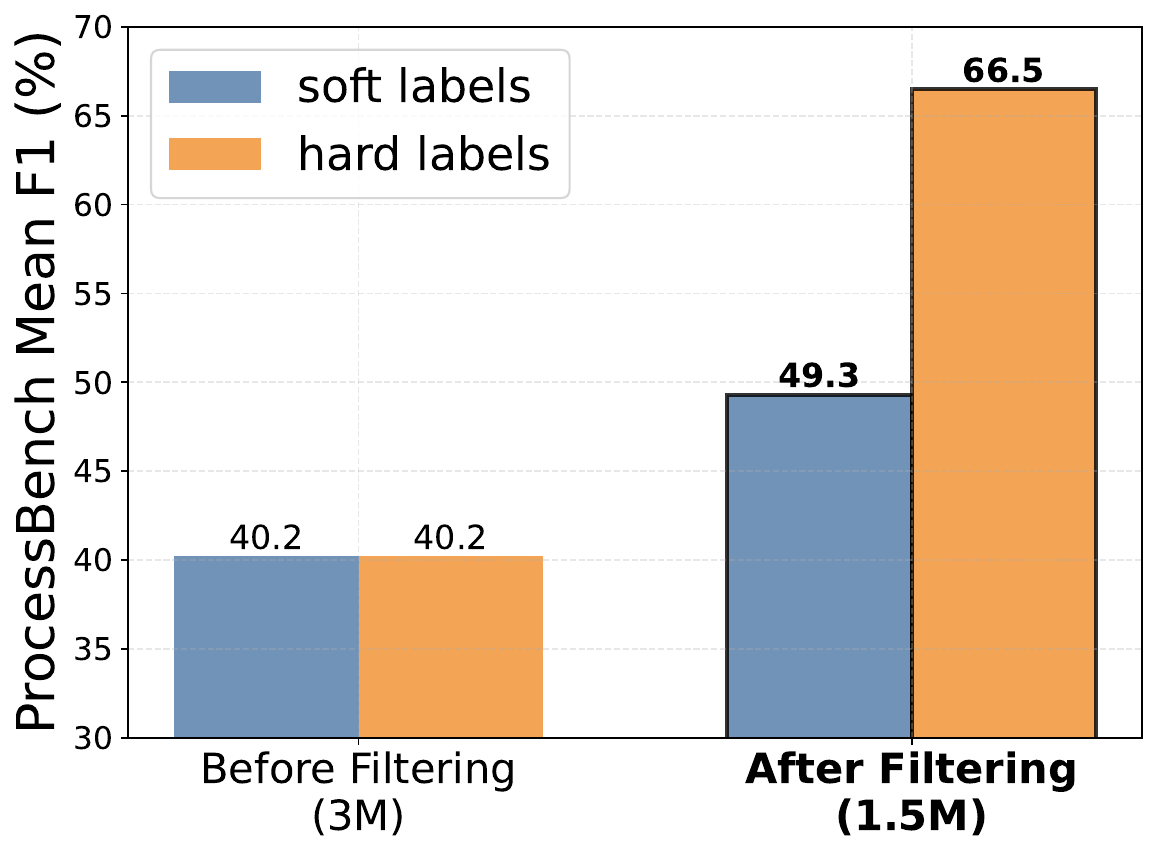}
        \caption{Performance comparison on \textsc{ProcessBench} for PRMs trained on soft and hard labels before and after consensus filtering.}
        \label{fig:soft_hard_process_bench}
    \end{minipage}
\end{figure}

Motivated by LLM-as-a-judge's encouraging results in Section \ref{sec:MC Estimation vs. LLM-as-a-judge vs. Human Annotation}, we naturally propose a simple yet efficient consensus Filtering mechanism that integrates LLM-as-a-judge with MC estimation. Based on the aforementioned 860K samples, the instances are only retained when both LLM-as-a-judge and MC estimation show consensus on the error reasoning step locations in the solution. As demonstrated in Figure \ref{fig:consensus_filtering}, it can be found that only approximately 40\% of the data are preserved after consensus filtering. For evaluation on \textsc{ProcessBench}, the results reveal that the reduced dataset after consensus filtering significantly outperforms MC estimation, and notably, achieves comparable performance to LLM-as-a-judge while using only 40\% of the data. Regarding the BoN evaluation, the performance variations among these three models are marginal. The limitations of BoN evaluation in PRMs will be elaborated on in Section \ref{sec:Bias in Best-of-N Sampling for PRM Performance Evaluation} later.

\subsubsection{Hard Label vs. Soft Label in MC Estimation}
Although we have previously demonstrated that MC estimation is not as effective as LLM-as-a-judge and human annotation, there remains a noteworthy point of MC estimation to be discussed, i.e., whether to train with soft label or hard label. We construct 3 million training data using MC estimation, where for each reasoning step we perform 8 completions. Subsequently, we apply the consensus filtering strategy discussed in Section \ref{sec:Stringent Data Filtering Mechanisms Required in MC Estimation} to filter the 3 million samples, which reduces the dataset to 1.5 million samples. We respectively train PRMs using both soft labels and hard labels on 3 million and 1.5 million data.

The performance of trained PRMs on Best-of-8 and \textsc{ProcessBench} are illustrated in Figure \ref{fig:soft_hard_bon} and \ref{fig:soft_hard_process_bench} separately. Before data filtering, the performance difference between soft and hard labels is not significant, which we attribute to the high noise level masking their distinctions. However, this difference becomes much more pronounced after data filtering, with hard labels substantially outperforming soft labels on both Best-of-8 and \textsc{ProcessBench}. We consider the limitations of soft labels are: (1) as discussed in Section \ref{sec:Distinguishing PRMs from Value Models}, the correctness of steps (i.e., rewards) should be deterministic. Training PRMs with soft labels that represent future possibilities introduces additional noise. For instance, when numerous completely correct steps are assigned with soft labels lower than 1, it actually reduces the model's ability to discriminate between positive and negative labels; (2) only 8 completions for step correctness estimation exhibit high variance and are relatively crude. Although we can achieve better estimation accuracy by increasing the number of completions, the associated costs may outweigh the incremental benefits. Moreover, the experimental results indicate that the consensus filtering strategy yields performance benefits across both soft and hard label schemes.

Last but not least, we investigate the threshold selection for distinguishing between positive and negative labels based on the MC estimation result of 8 completions. Following our previous experimental setup, we conduct a series of experiments on the 3 million with threshold values from 1/8 to 7/8 at 1/8 intervals, with results shown in Figure \ref{fig:threshold}. It can be easily observed that as the threshold increases, the performance deteriorates on both Best-of-8 and \textsc{ProcessBench}, indicating that using an MC estimated value of 0 as the negative label and all others as positive labels yields the best results. Therefore, if we have to rely on MC estimation for step-wise correctness verification, we suggest setting the threshold to 0, meaning that a step is considered correct if any completion start from this step reaches the correct final answer. This threshold has also been employed throughout our all experimental studies.

\subsubsection{Summary}
Through extensive experimentation, we have demonstrated that MC estimation yields inferior performance and generalization compared to both LLM-as-a-judge and human annotation. However, incorporating MC estimation with LLM-as-a-judge via a consensus filtering strategy leads to enhanced performance and improved data efficiency. Furthermore, optimal results are achieved when treating MC estimation values of 0 as negative labels and training with hard labels.

\subsection{Bias in BoN Sampling for PRM Performance Evaluation}\label{sec:Bias in Best-of-N Sampling for PRM Performance Evaluation}
Although BoN evaluations are commonly used in PRM optimization, their effectiveness as a sole optimization criterion is worth careful consideration due to potential limitations in performance assessment.

\begin{figure}[!t]
    \begin{minipage}[t]{0.32\textwidth}
    \centering
    \includegraphics[width=\textwidth]{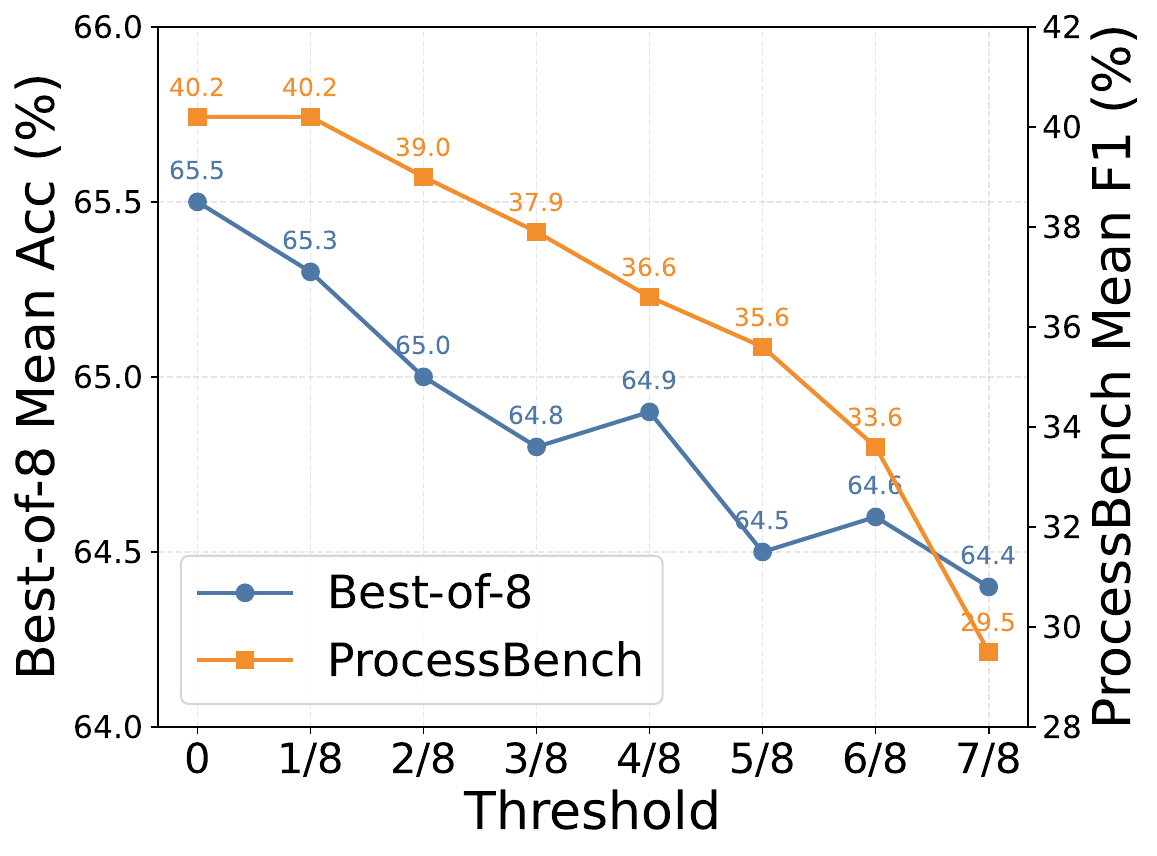}
    \caption{PRM Performance changes on Best-of-8 and \textsc{ProcessBench} across different hard label thresholds.}
    \label{fig:threshold}
    \end{minipage}
    \hfill
    \begin{minipage}[t]{0.32\textwidth}
        \centering
        \includegraphics[width=\textwidth]{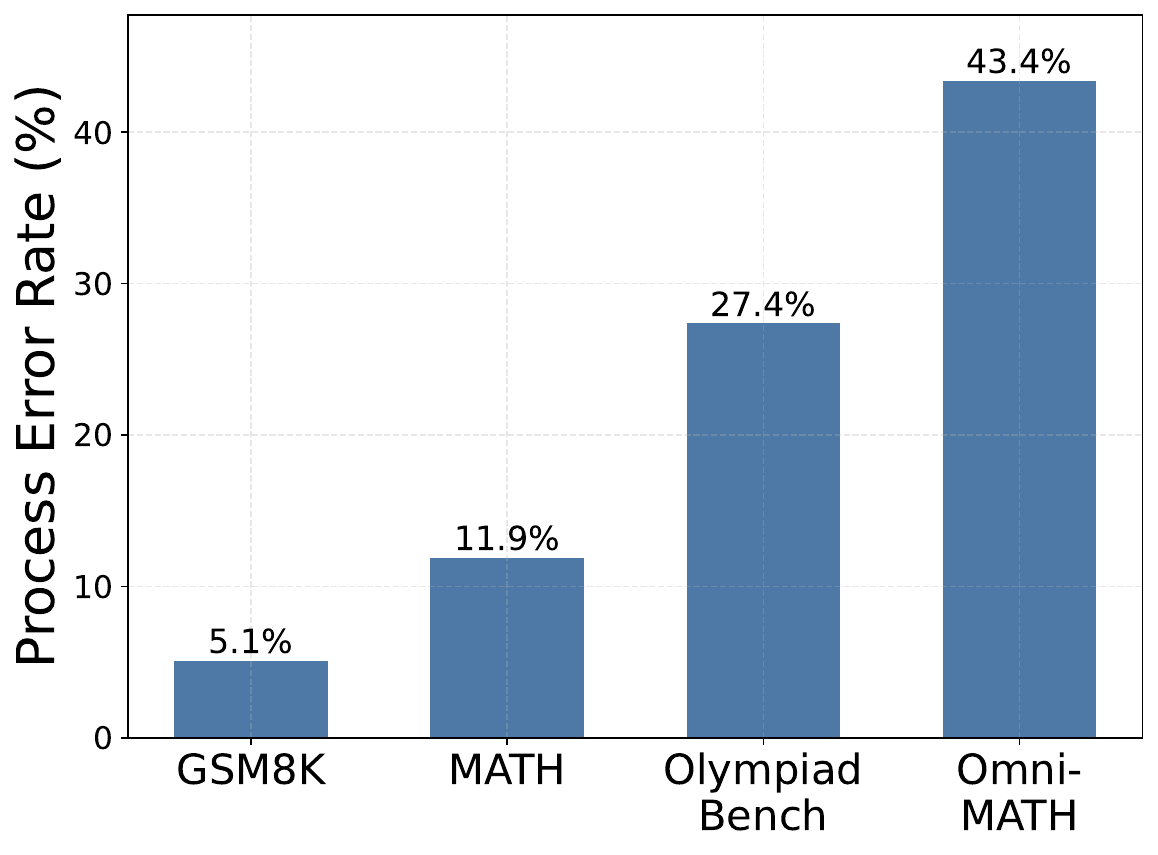}
        \caption{Proportion of cases where the policy model generates correct answers but incorrect reasoning steps.}
        \label{fig:process_error_rate}
    \end{minipage}
    \hfill
    \begin{minipage}[t]{0.32\textwidth}
      \centering
      \includegraphics[width=\textwidth]{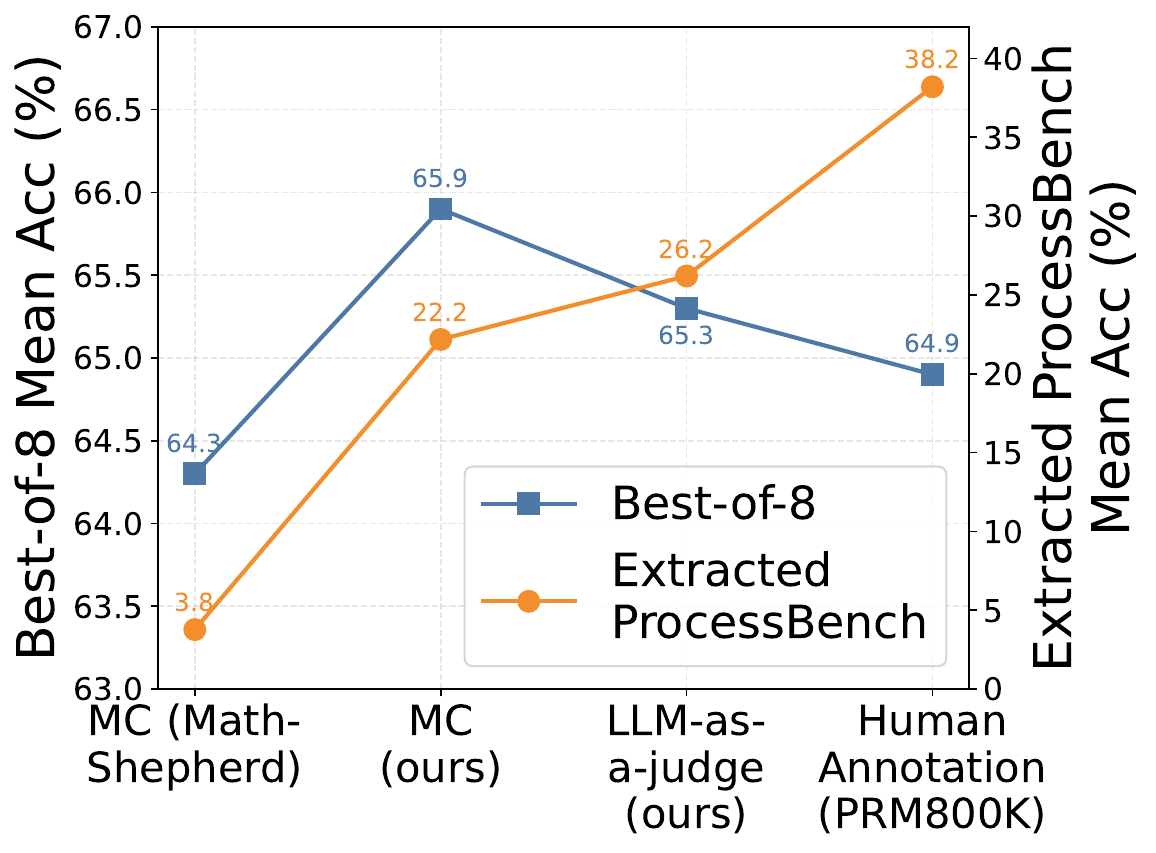}
    \caption{Performance trends on BoN and \textsc{ProcessBench} for models trained with different data sources.}
    \label{fig:trend}
    \end{minipage}
\end{figure}

\begin{table}[htp]
\small
\centering
\begin{tabular}{l cccc c}
\toprule
& GSM8K & MATH & OlympiadBench & Omni-MATH & Avg. \\
\midrule
\# samples & 7 & 94 & 161 & 259 &  \\
\midrule

\multicolumn{5}{l}{\textbf{1.5B}} \\
Skywork-PRM-1.5B              & 42.9  & 36.2 & 12.4           & 13.9      & 26.4 \\
\midrule
\multicolumn{5}{l}{\textbf{7B+}} \\
Math-Shepherd-PRM-7B          & 14.3  & 12.8 & 13.7           & 14.7      & 13.9 \\
RLHFlow-PRM-Mistral-8B        & 14.3  & 13.8 & 7.5            & 10.0      & 11.4 \\
RLHFlow-PRM-Deepseek-8B       & 0.0   & 18.1 & 9.9            & 10.8      & 9.7  \\
Skywork-PRM-7B                & \textbf{57.1}  & 26.6 & 14.3           & 13.1      & 27.8 \\
EurusPRM-Stage1              & 28.6 & 25.5  & 19.9  & 20.1 & 23.5 \\
EurusPRM-Stage2              & 42.9 &  27.7  & 18.0 &  20.8 & 27.4 \\
Qwen2.5-Math-7B-Math-Shepherd & 0.0   & 9.6  & 4.3            & 1.2       & 3.8  \\
Qwen2.5-Math-7B-PRM800K       & 42.9  & 50.0 & 31.7           & 28.2      & 38.2 \\
$\bigstar$ Qwen2.5-Math-PRM-7B           & 42.9  & \textbf{68.1} & \textbf{48.4}           & \textbf{56.0}      & \textbf{53.9} \\
\midrule
\multicolumn{5}{l}{\textbf{72B}} \\
$\bigstar$ Qwen2.5-Math-PRM-72B          & 28.6  & 76.6 & 62.7           & 64.5     & 58.1 \\ \bottomrule
\end{tabular}
\caption{The accuracy in identifying erroneous steps on the test cases of \textsc{ProcessBench} containing correct answers but erroneous reasoning steps. ``\# samples'' represents the number of test cases.}
\label{tab:correct answer with error process}
\end{table}

\subsubsection{Unreliable Policy Models Cause BoN-PRMs Misalignment}\label{sec:Unreliable Policy Models Cause BoN-PRMs Misalignment}
In an ideal scenario, the responses generated by the policy model would exhibit both correct answers and accurate solution steps or conversely, flawed processes would correspond to incorrect answers. However, existing policy models are prone to generating responses with correct answers but flawed processes, while BoN inherently only focuses on answers, leading to a misalignment between the evaluation criteria of BoN and the PRM objectives of process verification. To provide empirical evidence for this phenomenon, we sample 8 responses per query from GSM8K, MATH, OlympiadBench, and Omni-MATH using the policy model Qwen2.5-Math-7B-Instruct. Then we randomly choose correct-answer responses from them and conduct thorough manual annotations. As detailed in Figure \ref{fig:process_error_rate}, a substantial percentage of responses contain process errors while maintaining correct answers. Notably, compared with easy task GSM8K and hard task Omni-MATH, this phenomenon becomes more pronounced as the problem's complexity increases. This implies that an effective PRM might assign low scores to responses with correct answers but flawed processes, resulting in overall lower performance on the BoN evaluation.

\subsubsection{Limited Process Verification Capability in PRMs Lead to BoN Scores Inflation}
When the PRM cannot distinguish responses that have correct answers but flawed processes and assign them high scores, this leads to overestimated performance in the BoN evaluation, thereby creating an overly optimistic and potentially misleading assessment of PRM capabilities. 
To investigate the discriminative capability of PRMs for such cases, we extract instances from \textsc{ProcessBench} where answers are correct but processes are erroneous and analysis the detection accuracy of PRMs for these cases.
As shown in Figure \ref{fig:trend}, the PRMs trained on MC estimation, LLM-as-a-judge and human annotation exhibit completely opposite performance trends in BoN and extracted \textsc{ProcessBench} evaluation.
It can be observed that the model trained on our MC estimated data shows limited process verification capability but inflated results on the BoN.

On the other hand, as shown in Table \ref{tab:correct answer with error process}, except our released PRMs Qwen2.5-Math-PRM-7B and Qwen2.5-Math-PRM-72B, all other open-sourced PRMs demonstrate detection accuracy rates below 50\%. 
This limited discriminative capability indicates that PRMs struggle to differentiate between genuinely correct responses and those with merely superficial answer correctness in BoN evaluations. 
Consequently, this implies that beyond BoN evaluation, supplementary benchmarks are necessary to assess the actual capability of PRMs, especially in detecting process errors.

\begin{figure}[htp]
    \begin{minipage}[t]{0.49\textwidth}
      \centering
      \includegraphics[width=\textwidth]{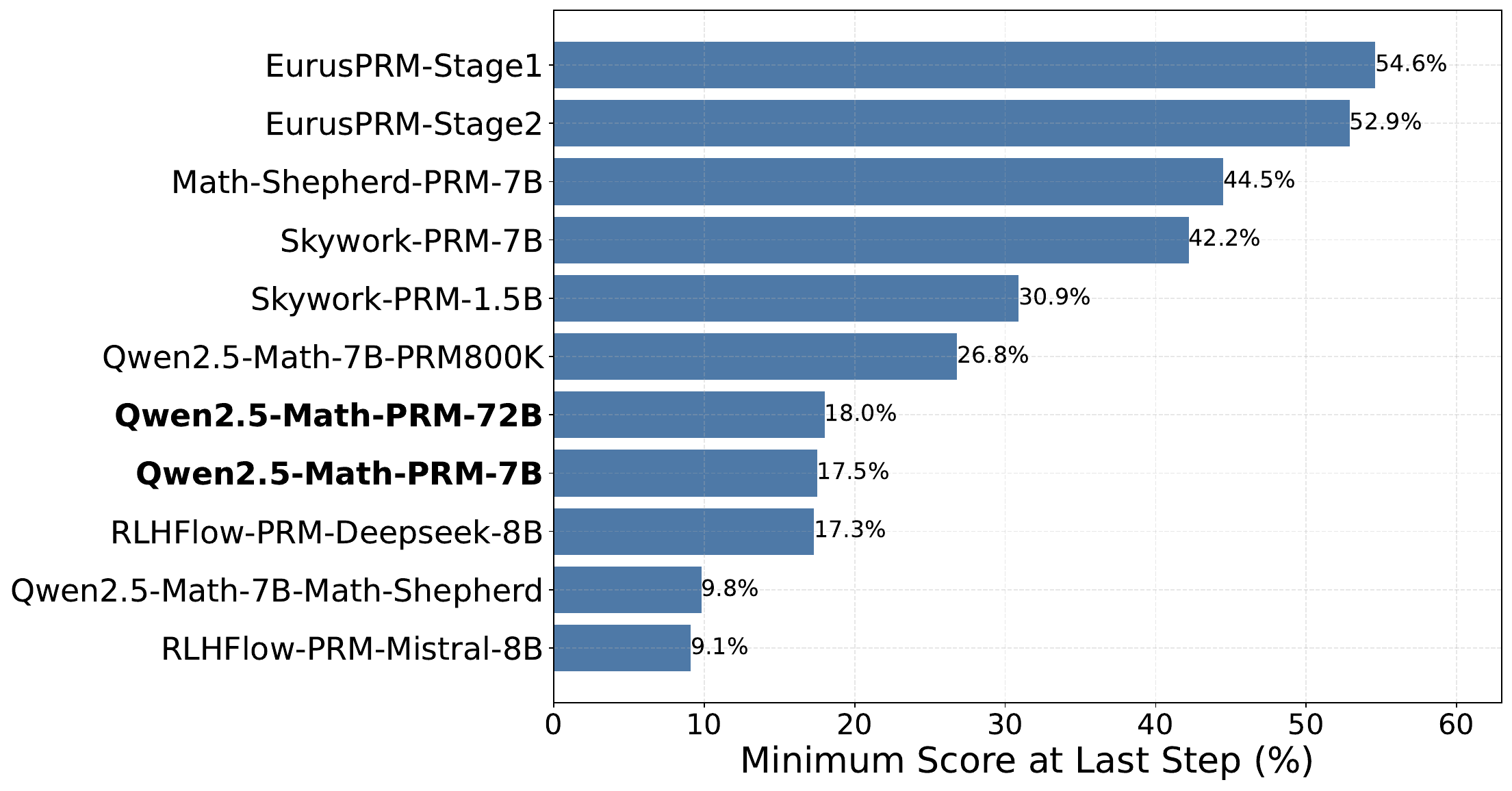}
    \caption{Percentage of responses where the minimum step score predict by PRMs appears in the final step (among all Best of 8 responses from Qwen2.5-Math-7B-Instruct).}
    \label{fig:minimum_score_at_last_step}
    \end{minipage}
    \hfill
    \begin{minipage}[t]{0.49\textwidth}
      \centering
      \includegraphics[width=\textwidth]{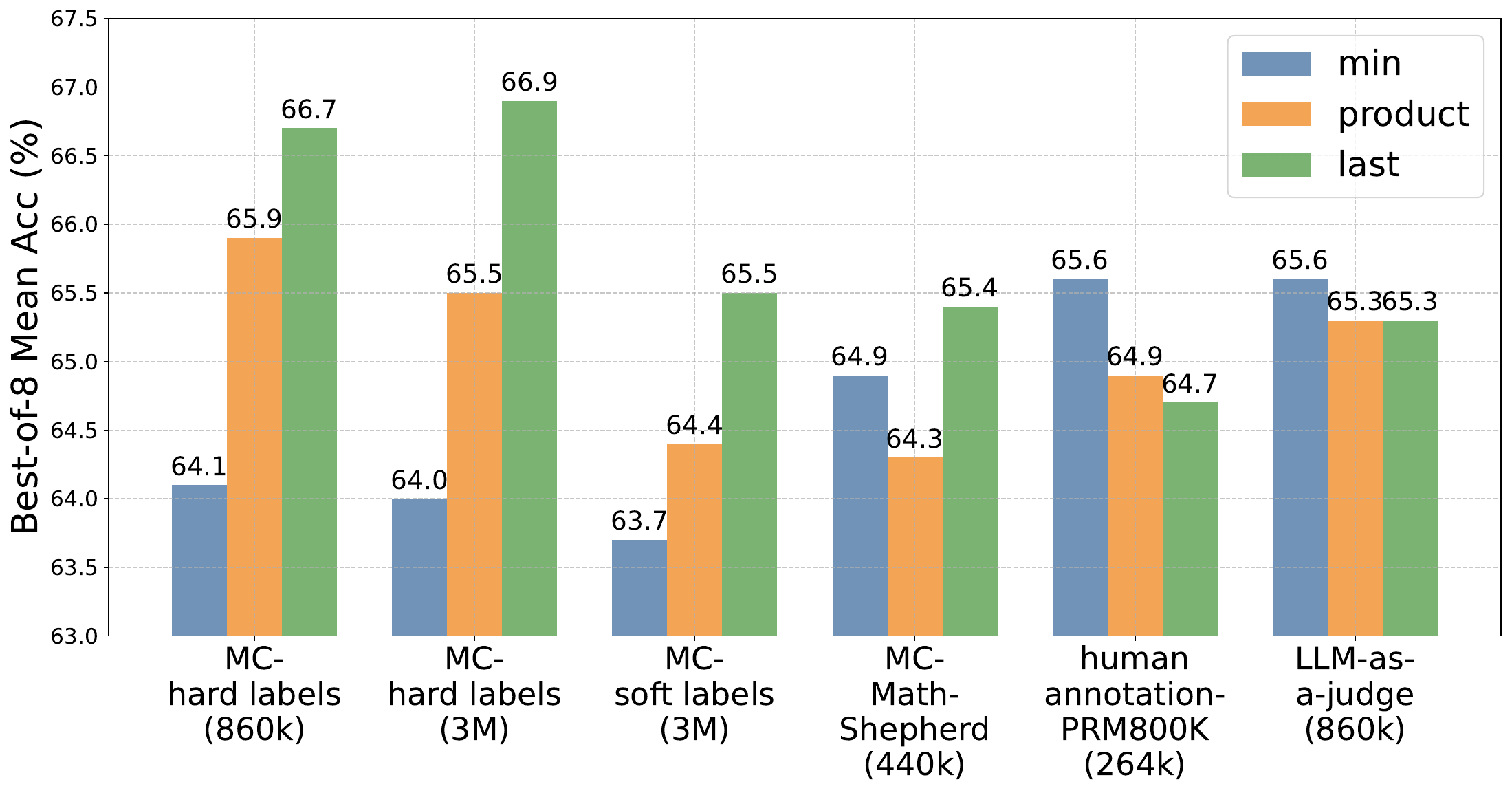}
    \caption{Performance on BoN across multiple PRMs with different scoring methods: minimum, product and last.}
    \label{fig:diffferent_scoring}
    \end{minipage}
\end{figure}

\subsubsection{Process-to-Outcome Shift in BoN Optimized PRMs}
The majority of current PRMs are optimized towards BoN. However, the limitations of BoN result in PRMs process-to-outcome shift. During the BoN selection process based on PRM-predicted scores and follow the scoring method for responses in \citep{prm}, it can be found that regardless of whether we employ the minimum score or the product of scores to evaluate the full solution, the lowest step score acts as the key limiting factor that affects the selection criteria of PRMs.

As shown in Figure \ref{fig:minimum_score_at_last_step}, we analyze the distribution of minimum step scores assigned by multiple open-sourced PRMs, specifically focusing on cases where the lowest score occurred at the final step, which typically contains the final answer. The results show that models EurusPRM-Stage1, EurusPRM-Stage2, Math-Shepherd-PRM-7B and Skywork-PRM-7B exhibit notably high proportions in this category, which exceed 40\%. In contrast, our released PRMs Qwen2.5-Math-PRM-72B and Qwen2.5-Math-PRM-7B exhibit a significantly lower proportion of minimum scores at the final step. 

This analysis reveals that some PRMs' performance in BoN evaluation is predominantly determined by final answer scores rather than intermediate reasoning steps, indicating a model degradation from process-based to outcome-oriented assessment. In other words, optimizing solely for the BoN evaluation has made current PRMs perform more like ORMs in practice. Hence, it is essential to supplement response-level evaluation BoN with step-level assessment methods to avoid the process-to-outcome shift. Specifically, we can employ process error localization tasks such as \textsc{ProcessBench}. Other commonly used step-wise BoN methodologies leverage the integration of PRMs or value models with search mechanisms, which provide a more granular assessment of process reliability. It worth noting that the latter requires more computational costs.

\subsubsection{Different PRMs, Different Optimal Scoring Strategies}
In the BoN evaluation, the overall solution score is derived by combining individual step scores. When each step's score represents the probability of that specific step being correct, it's generally acceptable to combine these step-level scores (through methods like product or minimum) to calculate the overall solution score. However, the situation becomes different when using MC estimation. In this case, each step's score actually estimates the probability of reaching the correct final answer in the future from the current position. Given this forward-looking nature of MC estimation, we should neither multiply the estimated probabilities across steps (as these estimates are dependent on each other), nor simply take the minimum estimated value from a particular step as the overall score. Instead, the estimated value from the final step naturally integrates information from the entire solution process, making it more suitable as the final score for the complete solution.

To validate that, we evaluate BoN in different scoring strategies for the PRMs trained on MC estimation, LLM-as-a-judge, and human annotation data, as shown in Figure \ref{fig:diffferent_scoring}. We found that in MC estimation, using the last score shows significantly better performance than product and minimum approaches across multiple PRMs. And the trend is the opposite for human annotation and LLM-as-a-judge. This suggests that if the PRM has to be trained via MC estimation and evaluated in BoN, the last score strategy may be more reasonable and effective. However, it's worth noting that this use of PRM in BoN has deviated from PRM's original intended purpose.

\subsubsection{Summary}
The above observations underscore critical limitations in BoN evaluation. \textit{Firstly}, the unreliable policy models generate responses with correct answers but flawed processes, leading to a misalignment between the evaluation criteria of BoN and the PRM objectives of process verification. \textit{Secondly}, the limited process verification capability makes PRMs demonstrate tolerance for the responses with correct answers but flawed reasoning processes, resulting in inflated BoN performance. \textit{Thirdly}, model optimization solely focused on BoN evaluation leads PRMs to drift to prioritize final answers over reasoning processes. Therefore, we argue that supplementary step-level evaluation plays a crucial role in PRM evaluation. \textit{Finally}, In BoN, different PRMs have different optimal scoring strategies. The last score strategy may be more reasonable and effective for the PRM trained via MC estimation. In contrast, product and minimum scoring are more appropriate for LLM-as-judge and human annotation.

\section{Our PRMs}
This section presents our methodology for overcoming the previously discussed limitations and the details of our trained PRM achieving state-of-the-art performance. Additionally, we outline our experimental settings, and baseline models for comparison and evaluation results.

\subsection{Training Details}
The data construction procedure comprises two primary phases: data expansion and data filtering. In the expansion phase, we follow the MC estimation to construct data described in Section \ref{sec:Training Setup}. We employ hard labels, where a response is classified as negative only if none of the 8 completions achieves the correct final answer. In the subsequent filtering phase, we employ the LLM instantiated by Qwen2.5-Instruct-72B \citep{qwen2.5} to serve as a critic to verify the reasoning process for all responses step by step, i.e., LLM-as-a-judge. We implement a simple yet efficient consensus filtering mechanism by filtering out instances where there is a discrepancy between the LLM-annotated and MC-estimated process labels. This ensures the retained data maintains high quality and consistency in the reasoning process annotation. For the training task, we employ cross-entropy loss on the tokens at the end of each step to train the binary classification task. We trained both 7B and 72B-parameter PRMs, initialized with Qwen2.5-Math-7B-Instruct and Qwen2.5-Math-72B-Instruct respectively.

\subsection{Experimental Setup}
To validate the effectiveness of our trained PRM Qwen2.5-Math-PRM-7B and Qwen2.5-Math-PRM-72B, we respectively conduct the response-level BoN evaluation and the step-level process errors identification task \textsc{ProcessBench} \citep{processbench}. 

\paragraph{Best-of-N} We follow the experimental setting in Section \ref{sec:Evaluation Setup}. In rm@8, we evaluate Outcome Reward Models (ORMs) and Process Reward Models (PRMs). For ORMs, we introduce Qwen2.5-Math-RM-72B \citep{qwen2.5-math}, which assigns a single score to each complete response. For PRMs, we compute the product of each step score as the final response score. We compare with the following PRMs:
\begin{itemize}
\item \textbf{Math-Shepherd-PRM-7B} \citep{math-shepherd}: determining process labels for each step by estimating the empirical probability of reaching the correct final answer.
\item \textbf{RLHFlow-PRM-Mistral-8B} \& \textbf{RLHFlow-PRM-Deepseek-8B} \citep{rlhflow-prm}: two LLaMA-3.1-based PRMs that adopt Math-Shepherd's training methodology while implementing different solution generation models and optimization objectives.
\item \textbf{Skywork-PRM-1.5B} \& \textbf{Skywork-PRM-7B} \citep{skywork-prm}: two recently released Qwen2.5-Math-based PRMs by Skywork.
\item \textbf{EurusPRM-Stage1} \& \textbf{EurusPRM-Stage2} \citep{cui2024process}: two PRMs trained using Implicit PRM approach \citep{yuan2024implicitprm} with 7B parameters, which obtains process rewards replying on the ORM trained on the response-level labels. 
\item \textbf{Qwen2.5-Math-7B-Math-Shepherd} \& \textbf{Qwen2.5-Math-7B-PRM800K}: two additional PRMs our developed by fine-tuning Qwen2.5-Math-7B-Instruct separately on the PRM800K \citep{prm} and Math-Shepherd \citep{math-shepherd} opensource datasets.
\end{itemize}

\paragraph{\textsc{ProcessBench}} The compared PRMs are consistent with the previously mentioned PRMs. For the LLM prompted as Critic Models, i.e., LLM-as-a-judge, we compare with proprietary language models GPT-4o-0806 \citep{gpt4o} and o1-mini \citep{o1-mini}, open-source language models Llama-3.3-70B-Instruct \citep{llama3}, Qwen2.5-Math-72B-Instruct \citep{qwen2.5-math}, Qwen2.5-72B-Instruct \citep{qwen2.5} and QwQ-32B-Preview \citep{qwq-32b-preview}. We also decompose the N-step response trajectory into N separate instances to enable individual scoring by the ORM Qwen2.5-Math-RM-72B.

\begin{table}[t]
\centering
\begin{adjustbox}{width=\textwidth}
\begin{tabular}{lcccccccc}
\toprule

\textbf{Setting} & \textbf{GSM8K} & \textbf{MATH} & \textbf{\makecell{Minerva\\Math}} & \textbf{\makecell{GaoKao\\2023 En}} & \textbf{\makecell{Olympiad\\Bench}} & \textbf{\makecell{College\\Math}} & \textbf{\makecell{MMLU\\STEM}} & \textbf{Avg.} \\ \midrule
pass@8 (Upper Bound) & 98.1 & 92 & 49.3 & 80.5 & 59.6 & 52.6 & 90.5 & 74.7 \\
maj@8 & 96.7 & 87.1 & 41.2 & 72.5 & 44.4 & 47.8 & 73.8 & 66.2 \\ \midrule
\textbf{\textsc{1.5B}} & & & & & & & \\
Skywork-PRM-1.5B              & 96.9  & 86.7 & 37.9         & 70.1           & 42.1           & 47.9         & 67.9      & 64.2 \\
\noalign{\vspace{4pt}}
\hdashline
\noalign{\vspace{4pt}}
\textbf{\textsc{7B+}} & & & & & & & \\
Math-Shepherd-PRM-7B          & \textbf{97.3}  & 85.4 & 37.9         & 70.6           & 40.4           & 47.2         & 70.5      & 64.2 \\
RLHFlow-PRM-Mistral-8B        & 97.0  & 86.1 & 37.1         & 70.6           & 41.2           & 47.6         & 69.5      & 64.2 \\
RLHFlow-PRM-Deepseek-8B       & \textbf{97.3}  & 86.3 & 40.8         & 70.9           & 42.2           & 47.2         & 69.3      & 64.9 \\
Skywork-PRM-7B                & \textbf{97.3}  & 87.3 & 38.2         & 71.9           & 43.7           & 47.8         & 67.7      & 64.8 \\
EurusPRM-Stage1 &  95.6	& 83.0 & 35.7 & 66.2 & 38.2 & 46.2 & 66.6 & 61.6 \\
EurusPRM-Stage2 &  95.4	& 83.4 & 34.9 & 67.3 & 39.1 & 46.3 & 67.3 & 62.0 \\
Qwen2.5-Math-7B-Math-Shepherd & 96.9 & 86.5 & 36.8 & 71.4 & 41.6 & 47.7 & 69.3 & 64.3 \\
Qwen2.5-Math-7B-PRM800K & 96.9 & 86.9 & 37.1 & 71.2 & 44.0 & 47.6 & 70.9 & 64.9 \\
$\bigstar$ Qwen2.5-Math-PRM-7B & 97.1 & \textbf{88.0} & \textbf{42.6} & \textbf{74.5} & \textbf{47.6} & \textbf{48.7} & \textbf{74.5} & \textbf{67.6} \\
\noalign{\vspace{4pt}}
\hdashline
\noalign{\vspace{4pt}}
\textbf{\textsc{72B}} & & & & & & & \\
Qwen2.5-Math-RM-72B & \textbf{97.9} & 88.5 & 42.6 & \textbf{75.1} & \textbf{49.9} & \textbf{49.6} & 78.7 & 68.9 \\
$\bigstar$ Qwen2.5-Math-PRM-72B          & 97.6  & \textbf{88.7} & \textbf{46.0}         & 74.3           & 48.1           & 49.3         & \textbf{81.1}      & \textbf{69.3}  \\
\bottomrule
\end{tabular}
\end{adjustbox}
\caption{Performance comparison on the Best-of-8 strategy of the policy model Qwen2.5-Math-
7B-Instruct. $\bigstar$ represents the models we trained.}
\label{tab:best-of-n on Qwen2.5-Math-7b-Instruct}
\end{table}

\subsection{Experimental Results}
\paragraph{Best-of-N} The evaluation on policy model Qwen2.5-Math-7b-Instruct is shown in Table \ref{tab:best-of-n on Qwen2.5-Math-7b-Instruct}. Qwen2.5-Math-PRM-7B demonstrates superior performance compared to other PRMs of equivalent model scale. Notably, it outperforms maj@8 across all 7 tasks, achieving an average improvement of 1.4\%. Furthermore, the Qwen2.5-Math-PRM-72B exhibits slightly better overall performance than Qwen2.5-Math-RM-72B, with particularly significant improvements observed in the Minerva Math and MMLU STEM tasks. Finally, Supplementary BoN results, including BoN performance on Policy model Qwen2.5-Math-72b-Instruct, alternative scoring strategies, evaluations on Chinese benchmarks, BoN with larger N values and BoN with LLM-as-a-judge are comprehensively documented in the Appendix \ref{Supplementary Experiments Results}.

\begin{table}[t]
\centering
\resizebox{\textwidth}{!}{
\begin{tabular}{lccccccccccccc}
\toprule
\multirow{2}{*}{\textbf{Model}} & \multicolumn{3}{c}{\textbf{GSM8K}} & \multicolumn{3}{c}{\textbf{MATH}} & \multicolumn{3}{c}{\textbf{OlympiadBench}} & \multicolumn{3}{c}{\textbf{Omni-MATH}} & \multirow{2}{*}{\textbf{Avg. F1}} \\
\cmidrule(lr){2-4} \cmidrule(lr){5-7} \cmidrule(lr){8-10} \cmidrule(lr){11-13}
 & error & correct & \textbf{F1} & error & correct  & \textbf{F1} & error & correct  & \textbf{F1} & error & correct  & \textbf{F1} \\
\midrule
\multicolumn{14}{l}{\textbf{\textit{LLM-as-judge, Proprietary language models}}} \\
GPT-4-0806 & 70.0 & 91.2 & 79.2 & 54.4 & 76.6 & 63.6 & 45.8 & 58.4 & 51.4 & 45.2 & 65.6 & 53.5 & 61.9 \\
o1-mini & 88.9 & 97.9 & \textbf{93.2} & 83.5 & 95.1 & \textbf{88.9} & 80.2 & 95.6 & \textbf{87.2} & 74.8 & 91.7 & \textbf{82.4} & \textbf{87.9} \\
\midrule
\multicolumn{14}{l}{\textbf{\textit{LLM-as-judge, Open-source language models}}} \\
Llama-3.3-70B-Instruct & 72.5 & 96.9 & 82.9 & 43.3 & 83.2 & 59.4 & 31.0 & 94.1 & 46.7 & 28.2 & 90.5 & 43.0 & 58.0\\
Qwen2.5-Math-72B-Instruct & 49.8 & 96.9 & 65.8 & 36.0 & 94.3 & 52.1 & 19.5 & 97.3 & 32.5 & 19.0 & 96.3 & 31.7 & 45.5\\
Qwen2.5-72B-Instruct & 62.8 & 96.9 & 76.2 & 46.3 & 93.1 & 61.8 & 38.7 & 92.6 & 54.6 & 36.6 & 90.9 & 52.2 & 61.2 \\
QwQ-32B-Preview & 81.6 & 95.3 & \textbf{88.0} & 78.1 & 79.3 & \textbf{78.7} & 61.4 & 54.6 & \textbf{57.8} & 55.7 & 68.0 & \textbf{61.3} & \textbf{71.5}\\
\midrule
\multicolumn{14}{l}{\textbf{\textit{PRMs}}} \\
\noalign{\vspace{4pt}}
\hdashline
\noalign{\vspace{4pt}}
\multicolumn{14}{l}{\textbf{1.5B}} \\
Skywork-PRM-1.5B & 50.2 & 71.5 & 59.0 & 37.9 & 65.2 & 48.0 & 15.4 & 26.0 & 19.3 & 13.6 & 32.8 & 19.2 & 36.4 \\
\noalign{\vspace{4pt}}
\hdashline
\noalign{\vspace{4pt}}
\multicolumn{14}{l}{\textbf{7B+}} \\
Math-Shepherd-PRM-7B & 32.4 & 91.7 & 47.9 & 18.0 & 82.0 & 29.5 & 15.0 & 71.1 & 24.8 & 14.2 & 73.0 & 23.8 & 31.5\\
RLHFlow-PRM-Mistral-8B & 33.8 & 99.0 & 50.4 & 21.7 & 72.2 & 33.4 & 8.2 & 43.1 & 13.8 & 9.6 & 45.2 & 15.8 & 28.4 \\
RLHFlow-PRM-Deepseek-8B & 24.2 & 98.4 & 38.8 & 21.4 & 80.0 & 33.8 & 10.1 & 51.0 & 16.9 & 10.9 & 51.9 & 16.9 & 26.6 \\
Skywork-PRM-7B & 61.8 & 82.9 & 70.8 & 43.8 & 62.2 & 53.6 & 17.9 & 31.9 & 22.9 & 14.0 & 41.9 & 21.0 & 42.1 \\
EurusPRM-Stage1 & 46.9	& 42.0	& 44.3	& 33.3	& 38.2	& 35.6	& 23.9	& 19.8	& 21.7	& 21.9	& 24.5	& 23.1	& 31.2 \\
EurusPRM-Stage2 & 51.2	& 44.0	& 47.3	& 36.4	& 35.0	& 35.7 & 	25.7	& 18.0	& 21.2	& 23.1	& 19.1	& 20.9	& 31.3 \\
Qwen2.5-Math-7B-Math-Shepherd & 46.4	& 95.9	& 62.5	& 18.9	& 96.6	& 31.6	& 7.4	& 93.8 & 13.7	& 4.0	& 95.0 & 7.7	& 28.9 \\
Qwen2.5-Math-7B-PRM800K & 53.1 & 95.3 & 68.2 & 48.0 & 90.1 & 62.6 & 35.7 & 87.3 & 50.7 & 29.8 & 86.1 & 44.3 & 56.5 \\
$\bigstar$ Qwen2.5-Math-PRM-7B &  72.0 & 	96.4	&  \textbf{82.4} & 	68.0	& 90.4	& \textbf{77.6}	& 55.7	& 85.5	& \textbf{67.5}	& 55.2	& 83.0	& \textbf{66.3}	& \textbf{73.5} \\
\noalign{\vspace{4pt}}
\hdashline
\noalign{\vspace{4pt}}
\multicolumn{14}{l}{\textbf{72B}} \\
Qwen2.5-Math-RM-72B & 41.1 & 46.1 & 43.5 & 39.7 & 58.1 & 47.2 & 28.1 & 56.6 & 37.6 & 18.8 & 50.2 & 27.4 & 38.9 \\
$\bigstar$ Qwen2.5-Math-PRM-72B & 78.7	& 97.9	& \textbf{87.3}	& 74.2	& 88.2	& \textbf{80.6}	& 67.9	& 82.0	& \textbf{74.3}	& 64.8	& 78.8	& \textbf{71.1}	& \textbf{78.3} \\
\bottomrule
\end{tabular}
}
\caption{Performance comparison on \textsc{ProcessBench}. $\bigstar$ represents the models we trained. We report the results in the same calculation method with \textsc{ProcessBench}.}
\label{tab:processbench}
\end{table}

\paragraph{\textsc{ProcessBench}} The evaluation results are presented in Table \ref{tab:processbench}. When compared with LLM-as-judge, Qwen2.5-Math-PRM-7B in smaller model size demonstrates superior performance over all open-source models. For proprietary language models, Qwen2.5-Math-PRM-7B outperforms GPT-4o-0806, while there remains a performance gap compared to o1-mini. Furthermore, comparing with existing PRMs, both Qwen2.5-Math-PRM-7B and 72B exhibit substantial advantages over their counterparts. An interesting observation worth noting is that the ORM Qwen2.5-Math-RM-72B exhibits considerable capability in identifying step errors, even surpassing some open-source PRMs, which validates its potential as a complementary reward beyond solely rule-based mechanism.

\section{Related Work}
\paragraph{Reward Model in Mathematical Reasoning} To further improve mathematical reasoning accuracy, the reward model plays a crucial role in selecting the best answers. Two main types of reward models have emerged: (1) Outcome Reward Model (ORM) which provides an evaluation score for the entire solution, especially for the final answer. (2) Process Reward Model (PRM) \citep{uesato2022solvingmathwordproblems,prm} which evaluates each step in the reasoning process. Previous work \citep{prm,math-shepherd} has demonstrated that PRM outperforms ORM which exhibits greater potential, though it requires more high-quality training data.

\paragraph{Mathematical Reasoning Step Verification} There are two primary approaches to evaluating the correctness of reasoning steps. The first approach relies on human annotation \citep{prm}, which produces high-quality data but suffers from substantial costs. The second approach, which has attracted considerable research attention, focuses on automated evaluation of reasoning step correctness. Current automated methods can be categorized into two main types: (1) backward-propagation based methods that infer step correctness from solution outcomes, including MC estimation \citep{math-shepherd,luo2024improvemathematicalreasoninglanguage,chen2024alphamathzeroprocesssupervision}, progressive ORM labeling \citep{r3}, and credit assignment \citep{wang2024qimprovingmultistepreasoning,cui2024process,yuan2024implicitprm} techniques; (2) prompting-based methods that leverage LLMs serve as critic, i.e., LLM-as-a-judge \citep{zhang2024generativeverifiersrewardmodeling,gao2024llmcriticshelpcatch,xia2024evaluatingmathematicalreasoningaccuracy} to assess step correctness directly. In this work, we integrate the two approaches MC estimation and LLM-as-a-judge.

\section{Conclusion}
In this paper, we investigate the Process Reward Model (PRM) and release an effective PRM that demonstrates superior performance. Firstly, we discuss the undesirable trials on MC estimation. Then we demonstrate that data construction via MC estimation yields inferior performance and generalization compared to both LLM-as-a-judge and human annotation through extensive experiments. Besides, we investigate the limitations of vanilla BoN evaluation for PRMs which leads to inaccurate assessment of the PRM's ability and causes an optimization bias that shifts focus from process-oriented to outcome-oriented verification. Finally, we propose a simple yet effective consensus filtering strategy combining MC estimation and LLM-as-a-judge to overcome the limitation of MC estimation. In terms of evaluation, we conduct the response-level BoN evaluation and the step-level process errors identification task \textsc{ProcessBench} to avoid the bias of relying solely on BoN. The experiments demonstrate our strategy significantly improves both data efficiency and model performance. In the future, there remains substantial potential in data construction and evaluation for PRMs, driving the development of more robust and reliable PRMs.

\paragraph{Limitation}
There are several limitations remained in our current work. Firstly, there exists a considerable performance gap between our PRMs and the BoN upper bound (pass@8), suggesting substantial optimization potential. 
Then the best practices for utilizing PRMs in reinforcement learning remain unexplored. 
Finally, although our approach combines LLM-as-a-judge with MC estimation for consensus filtering, the efficient utilization of existing high-quality human annotation data is still largely under-explored. For instance, gradually expanding high-quality datasets through weakly supervised methods can be investigated as a promising direction for future exploration.

\bibliographystyle{plainnat}
\bibliography{reference}

\appendix
\onecolumn

\section{PRM Guided Search}
We further integrate PRM with greedy search by generating N candidate steps at each step, evaluating these candidates using PRM scoring, and selecting the highest-scoring step for subsequent expansion. For the policy model, we employed Qwen2.5-7B-Instruct which has greater diversity in generation to sample 8 candidates at each step, with sampling parameters set to $temperature=1.0$ and $top\_p=1.0$. We conduct comparative experiments with ORM in BoN approach. As shown in Table \ref{tab:greedy search}, Qwen2.5-Math-PRM-72B with greedy search@8 is slightly superior performance compared to Qwen2.5-Math-RM-72B with orm@8. We argue the potentially smaller performance differential between PRM and ORM lies in the consistency of generated token counts between greedy search and BoN outputs. Furthermore, although greedy search always selects the highest-scoring candidate at each step, the highest-scoring step may not be the correct one. Therefore, implementing either Depth-First Search (DFS) with backtracking capabilities or search approaches incorporating score constraints could prove more suitable for this cases.

We choose the highest-scoring candidate at each step which the score predicted by PRM represents the correctness of this step. But such locally optimal choices may not lead to the correct final answer. In contrast, value models can predict the future probability of reaching the correct answer, rather than reflecting the correctness of the current step like rewards do, making them particularly well-suited for integration with search strategies. Based on these considerations, we believe there is still significant potential for exploration in the future regarding more appropriate search strategies or combining rewards and values to simultaneously consider both the correctness of the current step and the possibility of reaching the correct future outcomes.

\begin{table}[htp]
\centering
\resizebox{1\textwidth}{!}{
\begin{tabular}{lcccccccc}
\toprule
\textbf{Setting} & \textbf{GSM8K} & \textbf{MATH} & \textbf{\makecell{Minerva\\Math}} & \textbf{\makecell{GaoKao\\2023 En}} & \textbf{\makecell{Olympiad\\Bench}} & \textbf{\makecell{College\\Math}} & \textbf{\makecell{MMLU\\STEM}} & \textbf{Avg.} \\ \midrule
pass@8 (Upper Bound)                         & 96.9  & 89.6 & 48.2          & 79.7         & 58.4          & 55.0          & 81.6       & 72.8    \\
pass@1                          & 91.2  & 74.0 & 32.0          & 64.7         & 36.9          & 46.2          & 57.1       & 57.4    \\

maj@8                           & 93.7  & 80.3 & 37.1          & 69.9         & 45.8          & 48.5          & 61.9       & 62.5    \\
\midrule
\textbf{orm@8} & & & & & & & & \\
Qwen2.5-Math-RM-72B                         & 95.4  & 84.2 & \textbf{38.6}          & 73.0         & 48.6          & \textbf{50.1}          & \textbf{75.6}       & 66.5    \\
\midrule
\textbf{Greedy Search@8} & & & & & & & & \\
Skywork-PRM-7B & 95.3  & 83.2 & 33.8          & 70.4         & 44.1          & 48.2          & 60.1       & 62.2    \\
$\bigstar$ Qwen2.5-Math-PRM-7B              & 95.5  & 82.6 & 32.0          & 71.4         & 44.9          & 48.8          & 69.6       & 63.5    \\
$\bigstar$ Qwen2.5-Math-PRM-72B            & \textbf{95.9}  & \textbf{84.7} & 37.9          & \textbf{73.2}         & \textbf{48.9}          & 50.0          & 75.3       & \textbf{66.6}    \\
 \bottomrule
\end{tabular}
}
\caption{The performance of PRM guided greedy search and ORM of Best-of-8 with policy model Qwen2.5-7B-Instruct. For greedy search, 8 candidates is proposed at each step.}
\label{tab:greedy search}
\end{table}

\section{Supplementary BoN Results} \label{Supplementary Experiments Results}

\subsection{The BoN Evaluation on Qwen2.5-Math-72b-Instruct}
The BoN evaluation on policy model Qwen2.5-Math-72b-Instruct is shown in Table \ref{tab:best-of-n on Qwen2.5-Math-72b-Instruct}. Qwen2.5-
Math-7B-PRM outperforms other PRMs of equivalent model scale. However, its performance is inferior to maj@8, suggesting challenges in employing a 7B PRM for the supervision of 72B policy model-generated responses. Besides, Qwen2.5-Math-PRM-72B surpasses maj@8 in prm@8 and is comparable with Qwen2.5-Math-RM-72B in orm@8.

\begin{table}[htp]
\centering
\resizebox{1\textwidth}{!}{
\begin{tabular}{lcccccccc}
\toprule
\textbf{Setting} & \textbf{GSM8K} & \textbf{MATH} & \textbf{\makecell{Minerva\\Math}} & \textbf{\makecell{GaoKao\\2023 En}} & \textbf{\makecell{Olympiad\\Bench}} & \textbf{\makecell{College\\Math}} & \textbf{\makecell{MMLU\\STEM}} & \textbf{Avg.} \\ \midrule
pass@8 & 97.3 & 93.2 & 56.6 & 83.6 & 62.4 & 54.1 & 95.3 & 77.5 \\
maj@8 & 96.0 & 88.6 & 47.8 & 73.8 & 50.1 & 50.2 & 84.9 & 70.2 \\ \midrule
\textbf{\textsc{1.5B}} & & & & & & & \\
Skywork-PRM-1.5B              & 96.5  & 88.1 & 45.2         & 74.3           & 48.4           & 49.7         & 79.7      & 68.8 \\
\noalign{\vspace{4pt}}
\hdashline
\noalign{\vspace{4pt}}
\textbf{\textsc{7B+}} & & & & & & & \\
Math-Shepherd-PRM-7B          & 96.5  & 86.8 & 45.6         & 71.9           & 49.2           & 49.5         & 77.5      & 68.1 \\
RLHFlow-PRM-Mistral-8B        & 96.6  & 87.5 & 46.3         & 73.5           & 48.9           & 49.4         & \textbf{83.4}      & 69.4 \\
RLHFlow-PRM-Deepseek-8B       & 96.5  & 87.7 & 44.5         & 73.5           & 48.7           & 49.4         & 84.6      & 69.3 \\
Skywork-PRM-7B                & \textbf{97.0}  & 89.0 & 47.1         & 75.3           & 49.8           & 49.9         & 76.3      & 69.2 \\
EurusPRM-Stage1 & 95.4 & 85.6 & 44.1 & 72.5 & 46.5 & 49.2 & 80.3 & 67.7 \\
EurusPRM-Stage2 & 95.3 & 85.1 & 44.9 & 72.5 & 47.1 & 49.0 & 80.2 & 67.7 \\
Qwen2.5-Math-7B-Math-Shepherd & 96.9 & 88.5 & 46.0 & 75.8 & 49.9 & 49.5 & 79.7 & 69.5 \\
Qwen2.5-Math-7B-PRM800K & 96.5 & 88.9 & \textbf{47.4} & 75.3 & 50.7 & 50.1 & 76.6 & 69.4 \\
$\bigstar$ Qwen2.5-Math-PRM-7B & 96.8 & \textbf{89.6} & 46.7 & \textbf{77.7} & \textbf{51.4} & \textbf{50.4} & 76.4 & \textbf{69.9} \\ 
\noalign{\vspace{4pt}}
\hdashline
\noalign{\vspace{4pt}}
\textbf{\textsc{72B}} & & & & & & & \\
Qwen2.5-Math-RM-72B & 96.4 & 89.8 & \textbf{47.4} & 76.9 & \textbf{54.5} & \textbf{50.6} & 80.1 & \textbf{70.8} \\
$\bigstar$ Qwen2.5-Math-PRM-72B          & \textbf{96.4}  & \textbf{89.9} & 46.0         & \textbf{77.4}           & 52.9           & 50.1         & \textbf{82.3}      & 70.7 \\
\bottomrule
\end{tabular}
}
\caption{Performance comparison on the Best-of-8 strategy of the policy model Qwen2.5-Math-72B-Instruct. $\bigstar$ represents the models we trained.}
\label{tab:best-of-n on Qwen2.5-Math-72b-Instruct}
\end{table}

\subsection{The BoN Evaluation with Various Scoring Strategies}
We demonstrate experimental results using the last step score, the minimum step score or the production of step scores as the solution-level score. The BoN results with policy model Qwen2.5-Math-7B-Instruct and Qwen2.5-Math-72B-Instruct are shown in Table \ref{tab:bon results with different scoring on Qwen2.5-Math-7B-Instruct} and Table \ref{tab:bon results with different scoring on Qwen2.5-Math-72B-Instruct} respectively.

\subsection{The BoN Evaluation on Chinese Benchmarks}
We evaluate across three Chinese benchmarks including Chinese math benchmarks CMATH~\citep{cmath}, GaoKao Math Cloze~\citep{agieval}, and GaoKao Math QA~\citep{agieval} following \cite{qwen2.5-math}, as shown in Table \ref{tab:zh bon results with different scoring on Qwen2.5-Math-7B-Instruct} and Table \ref{tab:zh bon results with different scoring on Qwen2.5-Math-72B-Instruct}.

\subsection{BoN with Larger N Values}
To validate the effectiveness of our PRMs on the BoN with larger N values, we conduct additional Best-of-8 experiments on the policy model Qwen2.5-Math-7b-Instruct across diverse tasks including MATH500 \citep{prm}, AIME24~\footnote{\url{https://huggingface.co/datasets/AI-MO/aimo-validation-aime}}, AMC23~\footnote{\url{https://huggingface.co/datasets/AI-MO/aimo-validation-amc}}, Minerva Math \citep{minerva}, GaoKao 2023 En \citep{mario} and OlympiadBench \citep{olympiadbench}.
The results are presented in the Table \ref{tab:best-of-64 on Qwen2.5-Math-7b-Instruct} and it can be found that our PRMs maintain superior performance compared to other PRMs, especially on MATH500.

\begin{table}[htbp]
\centering
\resizebox{1\textwidth}{!}{
\begin{tabular}{lccccccc}
\toprule
\textbf{Setting} & \textbf{MATH500} & \textbf{AIME24} & \textbf{AMC23} & \textbf{\makecell{Minerva\\Math}} & \textbf{\makecell{GaoKao\\2023 En}} & \textbf{\makecell{Olympiad\\Bench}} & \textbf{Avg.} \\ \midrule
pass@64 & 96.0 & 50.0 & 95.0 & 56.6 & 86.8 & 73.5 & 76.3 \\
maj@64 & 84.2 & 16.7 & 77.5 & 34.6 & 73.8 & 51.1 & 56.3 \\
\midrule

\multicolumn{8}{l}{\textbf{1.5B}} \\
Skywork-PRM-1.5B & 81.2 & 20.0 & 62.5 & 31.6 & 70.9 & 46.5 & 52.1 \\
\noalign{\vspace{4pt}}
\hdashline
\noalign{\vspace{4pt}}
\multicolumn{8}{l}{\textbf{7B+}} \\
Math-Shepherd-PRM-7B & 79.6 & 20.0 & 62.5 & 32.4 & 70.1 & 43.9 & 51.4 \\
RLHFlow-PRM-Mistral-8B & 82.4 & 20.0 & 62.5 & 30.9 & 69.1 & 45.9 & 51.8 \\
RLHFlow-PRM-Deepseek-8B & 80.2 & 20.0 & \textbf{67.5} & \textbf{35.3} & 69.1 & 46.2 & 53.1 \\
Skywork-PRM-7B & 84.6 & 20.0 & \textbf{67.5} & 32.0 & 71.2 & 47.1 & 53.7 \\
EurusPRM-Stage1 & 76.0 & 10.0 & 55.0 & 27.6 & 66.5 & 40.0 & 45.9 \\
EurusPRM-Stage2 & 76.2 & 10.0 & 52.5 & 27.9 & 67.0 & 40.3 & 45.7 \\
Qwen2.5-Math-7B-Math-Shepherd & 84.2 & \textbf{23.3} & \textbf{67.5} & 34.6 & 72.5 & 47.4 & 54.9 \\
Qwen2.5-Math-7B-PRM800K & 83.6 & \textbf{23.3} & \textbf{67.5} & 33.8 & 74.8 & 48.3 & 55.2 \\
$\bigstar$ Qwen2.5-Math-PRM-7B  & \textbf{87.8} & 20.0 & \textbf{67.5} & 33.8 & \textbf{75.8} & \textbf{51.4} & \textbf{56.1} \\
\noalign{\vspace{4pt}}
\hdashline
\noalign{\vspace{4pt}}
\multicolumn{8}{l}{\textbf{72B}} \\
Qwen2.5-Math-RM-72B & 82.0 & \textbf{36.7} & \textbf{75.0} & 37.5 & \textbf{77.7} & 54.1 & \textbf{60.5} \\
$\bigstar$ Qwen2.5-Math-PRM-72B  & \textbf{87.8} & 23.3 & 72.5 & \textbf{38.6} & 77.4 & \textbf{55.3} & 59.2 \\
\bottomrule
\end{tabular}
}
\caption{Performance comparison on the Best-of-64 strategy of the policy model Qwen2.5-Math-7B-Instruct. $\bigstar$ represents the models we trained.}
\label{tab:best-of-64 on Qwen2.5-Math-7b-Instruct}
\end{table}

\subsection{Best-of-8 with LLM-as-a-judge}
Regarding BoN evaluation with LLMs, there are two ways to implement: pairwise and pointwise. For pairwise comparison, we employ a single-elimination tournament method. For N responses, we conduct N-1 comparisons to determine the optimal response. In terms of pointwise comparison, we score each step 1 for correct and 0 for incorrect. We then calculate the proportion of correct steps across all steps and select the response with the highest percentage of correct steps as the best response. The experiment are conduct on the policy model Qwen2.5-Math-7B-Instruct and Qwen2.5-Math-72B-Instruct and the results are shown in Table \ref{tab:best-of-8 comparison with llm-as-a-judge on Qwen2.5-Math-7b-Instruct} and Table \ref{tab:best-of-8 comparison with llm-as-a-judge on Qwen2.5-Math-72b-Instruct} respectively.

\begin{table*}[htp]
\centering
\resizebox{\textwidth}{!}{
\begin{tabular}{lcccccccc}
\toprule

\textbf{Setting} & \textbf{GSM8K} & \textbf{MATH} & \textbf{\makecell{Minerva\\Math}} & \textbf{\makecell{GaoKao\\2023 En}} & \textbf{\makecell{Olympiad\\Bench}} & \textbf{\makecell{College\\Math}} & \textbf{\makecell{MMLU\\STEM}} & \textbf{Avg.} \\ \midrule
pass@8 (Upper Bound) & 98.1 & 92 & 49.3 & 80.5 & 59.6 & 52.6 & 90.5 & 74.7 \\
maj@8 & 96.7 & 87.1 & 41.2 & 72.5 & 44.4 & 47.8 & 73.8 & 66.2 \\ \midrule

\multicolumn{9}{l}{\textbf{\textit{LLM-as-a-judge, Open-source language models}}} \\
\textbf{\textsc{pointwise}} & & & & & & & \\
QwQ-32B-Preview & 97.0 & 86.0 & 39.3 & 70.1 & 46.2 & 47.9 & 70.5 & 65.3 \\
Qwen2.5-72B-Instruct & 97.0 & 85.6 & 40.1 & 70.9 & 43.4 & 47.9 & 73.4 & 65.5 \\
\noalign{\vspace{4pt}}
\hdashline
\noalign{\vspace{4pt}}
\textbf{\textsc{pairwise}} & & & & & & & \\
QwQ-32B-Preview & 97.6 & 89.2 & 40.8 & 75.8 & 50.4 & 48.9 & 70.5 & 67.6 \\
Qwen2.5-72B-Instruct & 97.3 & 86.8 & 40.8 & 73.5 & 45.0 & 48.4 & 74.5 & 66.6 \\
\midrule
\multicolumn{9}{l}{\textbf{\textit{PRMs}}} \\
Qwen2.5-Math-PRM-7B  & 97.1 & 88.0 & 42.6 & 74.5 & 47.6 & 48.7 & 74.5 & 67.6 \\
Qwen2.5-Math-PRM-72B           & 97.6  & 88.7 & 46.0         & 74.3           & 48.1           & 49.3         & 81.1      & 69.3  \\
\bottomrule
\end{tabular}
}
\caption{Performance comparison with LLM-as-a-judge on the Best-of-8 strategy of the policy model Qwen2.5-Math-7B-Instruct.}
\label{tab:best-of-8 comparison with llm-as-a-judge on Qwen2.5-Math-7b-Instruct}
\end{table*}

\begin{table*}[htp]
\centering
\resizebox{\textwidth}{!}{
\begin{tabular}{lcccccccc}
\toprule

\textbf{Setting} & \textbf{GSM8K} & \textbf{MATH} & \textbf{\makecell{Minerva\\Math}} & \textbf{\makecell{GaoKao\\2023 En}} & \textbf{\makecell{Olympiad\\Bench}} & \textbf{\makecell{College\\Math}} & \textbf{\makecell{MMLU\\STEM}} & \textbf{Avg.} \\ \midrule
pass@8 (Upper Bound) & 97.3 & 93.2 & 56.6 & 83.6 & 62.4 & 54.1 & 95.3 & 77.5 \\
maj@8 & 96.0 & 88.6 & 47.8 & 73.8 & 50.1 & 50.2 & 84.9 & 70.2 \\ \midrule

\multicolumn{9}{l}{\textbf{\textit{LLM-as-a-judge, Open-source language models}}} \\
\textbf{\textsc{pointwise}} & & & & & & & \\
QwQ-32B-Preview & 96.2 & 88.3 & 46.3 & 75.3 & 51.0 & 50.0 & 74.9 & 68.9 \\
Qwen2.5-72B-Instruct & 96.5 & 87.8 & 47.4 & 76.4 & 48.9 & 50.0 & 76.0 & 69.0 \\
\noalign{\vspace{4pt}}
\hdashline
\noalign{\vspace{4pt}}
\textbf{\textsc{pairwise}} & & & & & & & \\
QwQ-32B-Preview & 96.4 & 90.9 & 46.0 & 79.5 & 55.1 & 50.5 & 73.6 & 70.3 \\
Qwen2.5-72B-Instruct & 96.1 & 88.2 & 43.4 & 75.3 & 50.1 & 49.6 & 71.4 & 67.7 \\
\midrule
\multicolumn{9}{l}{\textbf{\textit{PRMs}}} \\
Qwen2.5-Math-PRM-7B  & 96.8 & 89.6 & 46.7 & 77.7 & 51.4 & 50.4 & 76.4 & 69.9 \\
Qwen2.5-Math-PRM-72B & 96.4 & 89.9 & 46.0 & 77.4 & 52.9 & 50.1 & 82.3 & 70.7 \\
\bottomrule
\end{tabular}
}
\caption{Performance comparison with LLM-as-a-judge on the Best-of-8 strategy of the policy model Qwen2.5-Math-72B-Instruct.}
\label{tab:best-of-8 comparison with llm-as-a-judge on Qwen2.5-Math-72b-Instruct}
\end{table*}

\section{Prompt Template for LLM-as-a-judge}
\label{sec:llm-as-a-judge prompt}
To construct PRM training data via LLM-as-a-judge, we use the following prompt.
\begin{tcolorbox}[title=Prompt for constructing PRM training data via LLM-as-a-judge,breakable]
\begin{verbatim}
I will provide a math problem along with a solution. They will be formatted as 
follows:

[Math Problem]

<math_problem>
...(math problem)...
</math_problem>

[Solution]

<paragraph_1>
...(paragraph 1 of solution)...
</paragraph_1>

...

<paragraph_n>
...(paragraph n of solution)...
</paragraph_n>

Your task is to review each paragraph of the solution in sequence, analyzing, 
verifying, and critiquing the reasoning in detail. You need to provide the 
analyses and the conclusion in the following format:

<analysis_1>
...(analysis of paragraph 1)...
</analysis_1>

...

<analysis_n>
...(analysis of paragraph n)...
</analysis_n>

<conclusion>
Correct/Incorrect
</conclusion>


* When you analyze each paragraph, you should use proper verification, 
recalculation, or reflection to indicate whether it is logically and 
mathematically valid. Please elaborate on the analysis process carefully.

* If an error is detected in any paragraph, you should describe the nature and 
cause of the error in detail, and suggest how to correct the error or the correct 
approach. Once a paragraph is found to contain any error, stop further analysis 
of subsequent paragraphs (as they may depend on the identified error) and directly 
provide the conclusion of "Incorrect."

For instance, given a solution of five paragraphs, if an error is found in the 
third paragraph, you should reply in the following format:

<analysis_1>
...(analysis of paragraph 1)...
</analysis_1>

<analysis_2>
...(analysis of paragraph 2)...
</analysis_3>

<analysis_3>
...(analysis of paragraph 3; since an error is found here, also provide detailed 
critique and correction guideline)...
</analysis_3>

<conclusion>
Incorrect
</conclusion>

Note that the analyses of paragraphs 4 and 5 should be skipped as the paragraph 
3 has been found to contain an error.

* Respond with your analyses and conclusion directly.

--------------------------------------------------

The following is the math problem and the solution for you task:

[Math Problem]

{tagged_problem}

[Solution]

{tagged_response}
\end{verbatim}

\end{tcolorbox}

\begin{table}[htp]
\centering
\resizebox{1\textwidth}{!}{
\begin{tabular}{lccccccccc}
\toprule

\textbf{Setting} & \textbf{Scoring} & \textbf{GSM8K} & \textbf{MATH} & \textbf{\makecell{Minerva\\Math}} & \textbf{\makecell{GaoKao\\2023 En}} & \textbf{\makecell{Olympiad\\Bench}} & \textbf{\makecell{College\\Math}} & \textbf{\makecell{MMLU\\STEM}} & \textbf{Avg.} \\ \midrule
pass@8 (Upper Bound) & - & 98.1 & 92 & 49.3 & 80.5 & 59.6 & 52.6 & 90.5 & 74.7 \\
maj@8 & - & 96.7 &  87.1 & 41.2 & 72.5 & 44.4 & 47.8 & 73.8 & 66.2 \\ \midrule
\multirow{3}{*}{Math-Shepherd-PRM-7B}          & last    & 96.8  & 85.2 & 39.0          & 70.1         & 42.8          & 47.2          & 67.7       & 64.1    \\
                                               & product & 97.3  & 85.4 & 37.9          & 70.6         & 40.4          & 47.2          & 70.5       & 64.2    \\
                                               & min     & 96.9  & 85.3 & 39.0          & 69.9         & 42.2          & 47.4          & 70.6       & 64.5    \\
                                               \midrule
\multirow{3}{*}{RLHFlow-PRM-Mistral-8B}        & last    & 97.0  & 85.3 & 39.0          & 71.2         & 44.0          & 47.1          & 64.0       & 63.9    \\
                                               & product & 97.0  & 86.1 & 37.1          & 70.6         & 41.2          & 47.6          & 69.5       & 64.2    \\
                                               & min     & 97.0  & 84.3 & 37.1          & 69.4         & 40.4          & 46.9          & 68.7       & 63.4    \\
                                               \midrule
\multirow{3}{*}{RLHFlow-PRM-Deepseek-8B}       & last    & 97.0  & 84.7 & 35.7          & 70.4         & 43.0          & 46.8          & 63.8       & 63.1    \\
                                               & product & 97.3  & 86.3 & 40.8          & 70.9         & 42.2          & 47.2          & 69.3       & 64.9    \\
                                               & min     & 97.3  & 84.5 & 38.2          & 69.6         & 40.7          & 46.5          & 67.6       & 63.5    \\
                                               \midrule
\multirow{3}{*}{Skywork-PRM-1.5B}              & last    & 96.8  & 86.4 & 39.0          & 71.7         & 45.0          & 47.9          & 68.2       & 65.0    \\
                                               & product & 96.9  & 86.7 & 37.9          & 70.1         & 42.1          & 47.9          & 67.9       & 64.2    \\
                                               & min     & 96.6  & 86.6 & 37.9          & 71.9         & 43.1          & 48.2          & 66.9       & 64.5    \\
                                               \midrule
\multirow{3}{*}{Skywork-PRM-7B}                & last    & 97.2  & 87.3 & 41.2          & 73.8         & 45.8          & 48.3          & 65.3       & 65.6    \\
                                               & product & 97.3  & 87.3 & 38.2          & 71.9         & 43.7          & 47.8          & 67.7       & 64.8    \\
                                               & min     & 96.7  & 87.0 & 39.7          & 71.2         & 42.5          & 48.2          & 66.6       & 64.6    \\
                                               \midrule
\multirow{3}{*}{EurusPRM-Stage1}               & last    & 94.7  & 79.7 & 32.7          & 61.6         & 33.8          & 45.7          & 63.4       & 58.8    \\
                                               & product & 95.6  & 83.0 & 35.7          & 66.2         & 38.2          & 46.2          & 66.6       & 61.6    \\
                                               & min     & 95.8  & 83.3 & 39.0          & 67.8         & 37.9          & 46.6          & 67.4       & 62.5    \\
                                               \midrule
\multirow{3}{*}{EurusPRM-Stage2}               & last    & 94.7  & 79.7 & 33.1          & 61.3         & 34.2          & 45.7          & 63.5       & 58.9    \\
                                               & product & 95.4  & 83.4 & 34.9          & 67.3         & 39.1          & 46.3          & 67.3       & 62.0    \\
                                               & min     & 96.1  & 83.6 & 39.3          & 68.8         & 38.8          & 46.7          & 67.5       & 63.0    \\
                                               \midrule
\multirow{3}{*}{Qwen2.5-Math-7B-Math-Shepherd} & last    & 97.1  & 87.7 & 38.6          & 73.8         & 44.6          & 48.1          & 68.0       & 65.4    \\
                                               & product & 96.9  & 86.5 & 36.8          & 71.4         & 41.6          & 47.7          & 69.3       & 64.3    \\
                                               & min     & 97.0  & 86.7 & 36.8          & 72.5         & 43.1          & 47.6          & 70.7       & 64.9    \\
                                               \midrule
\multirow{3}{*}{Qwen2.5-Math-7B-PRM800K}       & last    & 96.7  & 86.3 & 37.9          & 71.9         & 44.3          & 47.6          & 68.1       & 64.7    \\
                                               & product & 96.9  & 86.9 & 37.1          & 71.2         & 44.0          & 47.6          & 70.9       & 64.9    \\
                                               & min     & 96.9  & 86.6 & 39.7          & 71.7         & 45.6          & 47.8          & 71.1       & 65.6    \\
                                               \midrule
\multirow{3}{*}{$\bigstar$ Qwen2.5-Math-PRM-7B}           & last    & 96.9  & 87.2 & 39.0          & 73.5         & 45.5          & 48.5          & 72.0       & 66.1    \\
                                               & product & 97.1  & 88.0 & 42.6          & 74.5         & 47.6          & 48.7          & 74.5       & 67.6    \\
                                               & min     & 97.0  & 87.8 & 42.3          & 74.3         & 46.2          & 48.3          & 74.1       & 67.1    \\
                                               \midrule
\multirow{3}{*}{$\bigstar$ Qwen2.5-Math-PRM-72B}          & last    & 97.6  & 88.9 & 43.4          & 73.8         & 49.2          & 49.6          & 76.8       & 68.5    \\
                                               & product & 97.6  & 88.7 & 46.0          & 74.3         & 48.1          & 49.3          & 81.1       & 69.3    \\
                                               & min     & 97.6  & 88.8 & 45.2          & 74.5         & 48.1          & 49.2          & 80.9       & 69.2    \\ 
                                               \bottomrule
\end{tabular}
}
\caption{Performance comparison on the Best-of-8 strategy of the policy model Qwen2.5-Math-7B-Instruct with 3 scoring strategies: last, product and minimum. $\bigstar$ represents the models we trained.}
\label{tab:bon results with different scoring on Qwen2.5-Math-7B-Instruct}
\end{table}

\begin{table}[htp]
\centering
\resizebox{1\textwidth}{!}{
\begin{tabular}{lccccccccc}
\toprule
\textbf{Setting} & \textbf{Scoring} & \textbf{GSM8K} & \textbf{MATH} & \textbf{\makecell{Minerva\\Math}} & \textbf{\makecell{GaoKao\\2023 En}} & \textbf{\makecell{Olympiad\\Bench}} & \textbf{\makecell{College\\Math}} & \textbf{\makecell{MMLU\\STEM}} & \textbf{Avg.} \\ \midrule
pass@8 (Upper Bound) & - & 97.3 & 93.2 & 56.6 & 83.6 & 62.4 & 54.1 & 95.3 & 77.5 \\
maj@8 & - & 96.0 & 88.6 & 47.8 & 73.8 & 50.1 & 50.2 & 84.9 & 70.2 \\
\midrule
\multirow{3}{*}{Math-Shepherd-PRM-7B}          & last    & 96.2  & 87.0 & 46.7          & 73.0         & 47.3          & 49.8          & 76.3       & 68.0    \\
                                               & product & 96.5  & 86.8 & 45.6          & 71.9         & 49.2          & 49.5          & 77.5       & 68.1    \\
                                               & min     & 96.1  & 86.8 & 45.6          & 73.2         & 48.6          & 49.9          & 76.0       & 68.0    \\
                                               \midrule
\multirow{3}{*}{RLHFlow-PRM-Mistral-8B}        & last    & 96.3  & 86.6 & 44.9          & 74.3         & 47.6          & 49.3          & 67.1       & 66.6    \\
                                               & product & 96.6  & 87.5 & 46.3          & 73.5         & 48.9          & 49.4          & 83.4       & 69.4    \\
                                               & min     & 96.4  & 86.3 & 44.5          & 71.9         & 47.9          & 49.3          & 76.0       & 67.5    \\
                                               \midrule
\multirow{3}{*}{RLHFlow-PRM-Deepseek-8B}       & last    & 96.1  & 86.6 & 46.3          & 73.2         & 49.2          & 49.2          & 71.7       & 67.5    \\
                                               & product & 96.5  & 87.7 & 44.5          & 73.5         & 48.7          & 49.4          & 84.6       & 69.3    \\
                                               & min     & 96.6  & 87.4 & 44.1          & 74.0         & 48.6          & 49.3          & 74.8       & 67.8    \\
                                               \midrule
\multirow{3}{*}{Skywork-PRM-1.5B}              & last    & 96.1  & 88.6 & 44.9          & 72.2         & 47.9          & 50.1          & 74.2       & 67.7    \\
                                               & product & 96.5  & 88.1 & 45.2          & 74.3         & 48.4          & 49.7          & 79.7       & 68.8    \\
                                               & min     & 96.0  & 88.3 & 45.6          & 73.8         & 48.6          & 50.1          & 75.9       & 68.3    \\
                                               \midrule
\multirow{3}{*}{Skywork-PRM-7B}                & last    & 97.0  & 89.0 & 46.0          & 74.8         & 51.0          & 49.7          & 66.7       & 67.7    \\
                                               & product & 97.0  & 89.0 & 47.1          & 75.3         & 49.8          & 49.9          & 76.3       & 69.2    \\
                                               & min     & 96.9  & 89.2 & 46.7          & 73.5         & 49.8          & 49.8          & 73.2       & 68.4    \\
                                               \midrule
\multirow{3}{*}{EurusPRM-Stage1}               & last    & 95.9  & 87.3 & 44.9          & 72.7         & 47.0          & 49.4          & 78.4       & 67.9    \\
                                               & product & 95.4  & 85.6 & 44.1          & 72.5         & 46.5          & 49.2          & 80.3       & 67.7    \\
                                               & min     & 96.4  & 88.2 & 44.9          & 75.1         & 49.0          & 49.5          & 83.7       & 69.5    \\
                                               \midrule
\multirow{3}{*}{EurusPRM-Stage2}               & last    & 96.0  & 87.7 & 44.5          & 73.5         & 47.0          & 49.4          & 78.1       & 68.0    \\
                                               & product & 95.3  & 85.1 & 44.9          & 72.5         & 47.1          & 49.0          & 80.2       & 67.7    \\
                                               & min     & 96.5  & 88.6 & 45.2          & 75.3         & 48.9          & 49.6          & 83.3       & 69.6    \\
                                               \midrule
\multirow{3}{*}{Qwen2.5-Math-7B-Math-Shepherd} & last    & 97.0  & 89.6 & 44.9          & 77.4         & 50.8          & 50.5          & 74.9       & 69.3    \\
                                               & product & 96.9  & 88.5 & 46.0          & 75.8         & 49.9          & 49.5          & 79.7       & 69.5    \\
                                               & min     & 97.0  & 88.6 & 46.0          & 74.8         & 50.2          & 49.6          & 79.6       & 69.4    \\
                                               \midrule
\multirow{3}{*}{Qwen2.5-Math-7B-PRM800K}       & last    & 96.7  & 88.8 & 47.1          & 76.1         & 50.1          & 49.5          & 71.8       & 68.6    \\
                                               & product & 96.5  & 88.9 & 47.4          & 75.3         & 50.7          & 50.1          & 76.6       & 69.4    \\
                                               & min     & 96.5  & 89.1 & 47.1          & 76.1         & 50.7          & 49.9          & 75.3       & 69.2    \\
                                               \midrule
\multirow{3}{*}{$\bigstar$ Qwen2.5-Math-PRM-7B}           & last    & 96.8  & 89.0 & 46.7          & 75.3         & 49.8          & 50.3          & 78.4       & 69.5    \\
                                               & product & 96.8  & 89.6 & 46.7          & 77.7         & 51.4          & 50.4          & 76.4       & 69.9    \\
                                               & min     & 96.7  & 89.6 & 46.3          & 77.9         & 50.8          & 50.3          & 76.0       & 69.7    \\
                                               \midrule
\multirow{3}{*}{$\bigstar$ Qwen2.5-Math-PRM-72B}          & last    & 96.3  & 89.8 & 47.8          & 76.6         & 53.3          & 50.9          & 80.5       & 70.7    \\
                                               & product & 96.4  & 89.9 & 46.0          & 77.4         & 52.9          & 50.1          & 82.3       & 70.7    \\
                                               & min     & 96.4  & 89.7 & 46.3          & 77.7         & 52.4          & 50.4          & 81.2       & 70.6    \\ \bottomrule
\end{tabular}
}
\caption{Performance comparison on the Best-of-8 strategy of the policy model Qwen2.5-Math-72B-Instruct with 3 scoring strategies: last, product and minimum. $\bigstar$ represents the models we trained.}
\label{tab:bon results with different scoring on Qwen2.5-Math-72B-Instruct}
\end{table}

\begin{table}[htp]
\centering
\begin{tabular}{lccccc}
\toprule
\textbf{Setting} & \textbf{Scoring} & \textbf{CMATH} & \textbf{\makecell{CN Middle\\School 24}} & \textbf{GaoKao} & \textbf{Avg.}\\ \midrule
pass@8 (Upper Bound)               & -  & 95.3  & 82.2                 & 84.3        & 87.3    \\
maj@8                & -  & 92.7  & 78.2                 & 68.1        & 79.7    \\ \midrule
\multirow{3}{*}{Math-Shepherd-PRM-7B}          & last    & 91.8  & 80.2                 & 63.0        & 78.3    \\
                                               & product & 92.0  & 80.2                 & 69.1        & 80.4    \\
                                               & min     & 91.5  & 80.2                 & 69.8        & 80.5    \\
                                               \midrule
\multirow{3}{*}{RLHFlow-PRM-Mistral-8B}        & last    & 92.8  & 79.2                 & 57.2        & 76.4    \\
                                               & product & 92.7  & 77.2                 & 65.8        & 78.6    \\
                                               & min     & 92.8  & 76.2                 & 62.1        & 77.0    \\
                                               \midrule
\multirow{3}{*}{RLHFlow-PRM-Deepseek-8B}       & last    & 93.2  & 75.2                 & 56.9        & 75.1    \\
                                               & product & 92.7  & 76.2                 & 63.6        & 77.5    \\
                                               & min     & 93.0  & 74.3                 & 67.3        & 78.2    \\
                                               \midrule
\multirow{3}{*}{Skywork-PRM-1.5B}              & last    & 93.8  & 80.2                 & 66.6        & 80.2    \\
                                               & product & 92.8  & 79.2                 & 66.3        & 79.4    \\
                                               & min     & 93.3  & 80.2                 & 66.6        & 80.0    \\
                                               \midrule
\multirow{3}{*}{Skywork-PRM-7B}                & last    & 94.0  & 81.2                 & 66.7        & 80.6    \\
                                               & product & 93.3  & 79.2                 & 68.1        & 80.2    \\
                                               & min     & 93.8  & 80.2                 & 66.3        & 80.1    \\
                                               \midrule
\multirow{3}{*}{EurusPRM-Stage1}               & last    & 91.8  & 77.2                 & 55.4        & 74.8    \\
                                               & product & 91.7  & 77.2                 & 52.6        & 73.8    \\
                                               & min     & 91.7  & 78.2                 & 64.4        & 78.1    \\
                                               \midrule
\multirow{3}{*}{EurusPRM-Stage2}               & last    & 91.8  & 77.2                 & 55.7        & 74.9    \\
                                               & product & 92.0  & 77.2                 & 52.4        & 73.9    \\
                                               & min     & 92.0  & 78.2                 & 64.7        & 78.3    \\
                                               \midrule
\multirow{3}{*}{Qwen2.5-Math-7B-Math-Shepherd} & last    & 93.0  & 81.2                 & 65.4        & 79.9    \\
                                               & product & 93.0  & 79.2                 & 67.7        & 80.0    \\
                                               & min     & 92.5  & 80.2                 & 69.8        & 80.8    \\
                                               \midrule
\multirow{3}{*}{Qwen2.5-Math-7B-PRM800K}       & last    & 92.8  & 78.2                 & 67.1        & 79.4    \\
                                               & product & 92.7  & 77.2                 & 68.9        & 79.6    \\
                                               & min     & 93.0  & 77.2                 & 69.4        & 79.9    \\
                                               \midrule
\multirow{3}{*}{$\bigstar$ Qwen2.5-Math-PRM-7B}           & last    & 93.3  & 80.2                 & 68.2        & 80.6    \\
                                               & product & 93.7  & 80.2                 & 70.1        & 81.3    \\
                                               & min     & 93.5  & 80.2                 & 71.7        & 81.8    \\
                                               \midrule
\multirow{3}{*}{$\bigstar$ Qwen2.5-Math-PRM-72B}          & last    & 94.3  & 80.2                 & 72.1        & 82.2    \\
                                               & product & 94.2  & 80.2                 & 73.5        & 82.6    \\
                                               & min     & 94.2  & 80.2                 & 73.1        & 82.5    \\ \bottomrule
\end{tabular}
\caption{Best-of-8 performance comparison on the Chinese benchmarks with the policy model Qwen2.5-Math-7B-Instruct in 3 scoring strategies: last, product and minimum. $\bigstar$ represents the PRMs we trained.}
\label{tab:zh bon results with different scoring on Qwen2.5-Math-7B-Instruct}
\end{table}

\begin{table}[htp]
\centering
\begin{tabular}{lccccc}
\toprule
\textbf{Setting} & \textbf{Scoring} & \textbf{CMATH} & \textbf{\makecell{CN Middle\\School 24}} & \textbf{GaoKao} & \textbf{Avg.}\\ \midrule
pass@8 (Upper Bound)                                         &     -    & 96.8  & 83.2                 & 86.2        & 88.7    \\
maj@8                                          &     -    & 95.3  & 79.2                 & 75.0        & 83.2    \\ \midrule
\multirow{3}{*}{Math-Shepherd-PRM-7B}          & last    & 93.7  & 78.2                 & 73.2        & 81.7    \\
                                               & product & 94.0  & 80.2                 & 72.1        & 82.1    \\
                                               & min     & 93.5  & 80.2                 & 73.9        & 82.5    \\
                                               \midrule
\multirow{3}{*}{RLHFlow-PRM-Mistral-8B}        & last    & 94.3  & 79.2                 & 65.5        & 79.7    \\
                                               & product & 93.8  & 79.2                 & 72.0        & 81.7    \\
                                               & min     & 93.3  & 79.2                 & 71.2        & 81.2    \\
                                               \midrule
\multirow{3}{*}{RLHFlow-PRM-Deepseek-8B}       & last    & 94.3  & 79.2                 & 63.0        & 78.8    \\
                                               & product & 94.3  & 79.2                 & 72.5        & 82.0    \\
                                               & min     & 94.5  & 79.2                 & 73.5        & 82.4    \\
                                               \midrule
\multirow{3}{*}{Skywork-PRM-1.5B}              & last    & 94.8  & 80.2                 & 74.3        & 83.1    \\
                                               & product & 93.8  & 79.2                 & 69.7        & 80.9    \\
                                               & min     & 94.5  & 80.2                 & 74.6        & 83.1    \\
                                               \midrule
\multirow{3}{*}{Skywork-PRM-7B}                & last    & 95.3  & 80.2                 & 72.6        & 82.7    \\
                                               & product & 94.7  & 80.2                 & 71.5        & 82.1    \\
                                               & min     & 94.8  & 80.2                 & 76.0        & 83.7    \\
                                               \midrule
\multirow{3}{*}{EurusPRM-Stage1}               & last    & 94.0  & 79.2                 & 64.5        & 79.2    \\
                                               & product & 93.8  & 80.2                 & 64.5        & 79.5    \\
                                               & min     & 94.7  & 79.2                 & 70.8        & 81.6    \\
                                               \midrule
\multirow{3}{*}{EurusPRM-Stage2}               & last    & 94.2  & 79.2                 & 63.4        & 78.9    \\
                                               & product & 93.7  & 80.2                 & 65.4        & 79.8    \\
                                               & min     & 94.3  & 79.2                 & 69.7        & 81.1    \\
                                               \midrule
\multirow{3}{*}{Qwen2.5-Math-7B-Math-Shepherd} & last    & 95.0  & 81.2                 & 74.6        & 83.6    \\
                                               & product & 94.5  & 80.2                 & 73.0        & 82.6    \\
                                               & min     & 94.3  & 80.2                 & 71.5        & 82.0    \\
                                               \midrule
\multirow{3}{*}{Qwen2.5-Math-7B-PRM800K}       & last    & 94.2  & 79.2                 & 76.5        & 83.3    \\
                                               & product & 94.2  & 82.2                 & 70.8        & 82.4    \\
                                               & min     & 93.8  & 80.2                 & 72.9        & 82.3    \\
                                               \midrule
\multirow{3}{*}{$\bigstar$ Qwen2.5-Math-PRM-7B}                            & last    & 94.7  & 79.2                 & 74.5        & 82.8    \\
                                               & product & 94.3  & 81.2                 & 77.6        & 84.4    \\
                                               & min     & 94.5  & 81.2                 & 77.6        & 84.4    \\
                                               \midrule
\multirow{3}{*}{$\bigstar$ Qwen2.5-Math-PRM-72B}          & last    & 96.0  & 79.2                 & 76.1        & 83.8    \\
                                               & product & 96.0  & 80.2                 & 77.2        & 84.5    \\
                                               & min     & 95.8  & 80.2                 & 77.5        & 84.5    \\ \bottomrule
\end{tabular}
\caption{Best-of-8 performance comparison on the Chinese benchmarks with the policy model Qwen2.5-Math-72B-Instruct in 3 scoring strategies: last, product and minimum. $\bigstar$ represents the PRMs we trained.}
\label{tab:zh bon results with different scoring on Qwen2.5-Math-72B-Instruct}
\end{table}

\end{document}